\documentclass[runningheads]{llncs}

\usepackage{graphicx}
\usepackage{comment}
\usepackage{amsmath,amssymb}
\usepackage{color}
\usepackage{url}
\usepackage{hyperref}

\newif\ifreview
\reviewfalse

\ifreview
	\usepackage{lineno}

	\linenumbers
\fi

\usepackage{amsmath}
\usepackage{multirow}
\usepackage{booktabs}
\usepackage{amssymb}
\usepackage{pifont}
\usepackage{xcolor}
\usepackage{bbding}
\usepackage[misc]{ifsym}
\usepackage{stmaryrd}
\usepackage{subcaption}
\usepackage{color, colortbl}
\usepackage[utf8]{inputenc}
\usepackage[figuresright]{rotating}
\newcommand{\xmark}{\ding{55}}%

\usepackage{acro}
\usepackage{xspace}

\newcommand{\norm}[1]{\left\lVert#1\right\rVert}

\DeclareAcronym{mot}{
  short = MOT ,
  long = multiple object tracking,
}

\newcommand{\bdd}{BDD100K\xspace}
\newcommand{\shift}{SHIFT\xspace}

\newcommand{\mot}{\ac{mot}\xspace}

\makeatletter
\newcommand{\printfnsymbol}[1]{%
  \textsuperscript{\@fnsymbol{#1}}%
}
\makeatother

\begin{document}


\def\SubNumber{84}

\def\GCPRTrack{Fast Review Track}

\title{COOLer: Class-Incremental Learning for Appearance-Based Multiple Object Tracking}

\titlerunning{COOLer: Class-Incremental Learning for Multiple Object Tracking}

\ifreview
\else

	\author{Zhizheng Liu\thanks{Equal contribution.} \orcidID{0009-0006-9426-3718} \and
	Mattia Segu\printfnsymbol{1}\orcidID{0000-0002-9107-531X} \and \\
	Fisher Yu\inst{(}\textsuperscript{\Letter}\inst{)}\orcidID{0000-0001-8829-7344}}
	
	\authorrunning{Z. Liu et al.}
	
	\institute{ ETH Zürich,
8092 Zürich, Switzerland
 \\
	\email{\{liuzhi, segum\}@ethz.ch, i@yf.io}\\
	 }
\fi

\maketitle              

\begin{abstract}
%
%
Continual learning allows a model to learn multiple tasks sequentially while retaining the old knowledge without the training data of the preceding tasks. 
This paper extends the scope of continual learning research to class-incremental learning for \ac{mot}, which is desirable to accommodate the continuously evolving needs of autonomous systems.
Previous solutions for continual learning of object detectors do not address the data association stage of appearance-based trackers, leading to catastrophic forgetting of previous classes' re-identification features.
%
%
%
We introduce COOLer, a COntrastive- and cOntinual-Learning-based tracker, 
which incrementally learns to track new categories while preserving past knowledge 
by training on a combination of currently available ground truth labels and pseudo-labels generated by the past tracker. 
%
To further exacerbate the disentanglement of instance representations, we introduce a novel contrastive class-incremental instance representation learning technique.
%
%
Finally, we propose a practical evaluation protocol for continual learning for MOT and conduct experiments on the \bdd and \shift datasets. 
%
%
Experimental results demonstrate that COOLer continually learns while effectively addressing catastrophic forgetting of both tracking and detection. 
The code is available at \url{https://github.com/BoSmallEar/COOLer}.

%

\end{abstract}

\begin{figure}[t]
\centering
\includegraphics[width=1.0\columnwidth]{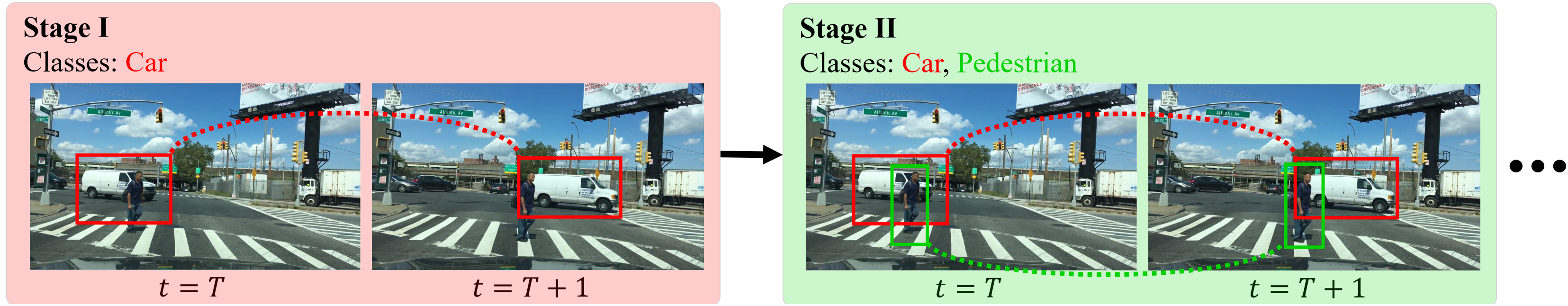}  
\caption{Illustration of the class-incremental learning problem for multiple object tracking. In a first stage, an MOT model can only track cars (red). When given annotations only for the novel class `pedestrian' (green), the objective is learning to track the new class without forgetting the previous one. }
\label{fig:teaser}
\end{figure}
\section{Introduction}
Continual learning aims at training a model to gradually extend its knowledge and learn multiple tasks sequentially without accessing the previous training data~\cite{chen2018lifelong}.
Since merely finetuning a pre-trained model on the new task would result in forgetting the knowledge learned from previous tasks - a problem known in literature as catastrophic forgetting~\cite{mccloskey1989catastrophic}- ad-hoc continual learning solutions are required.
As data distributions and practitioners' needs change over time, the practicality of continual learning has made it popular in recent years.

This paper addresses class-incremental learning for multiple object tracking (\mot), an important yet novel research problem that, to the best of our knowledge, has not been studied in previous literature.
%
%
%
%
%
%
%
\mot tracks multiple objects simultaneously from a video sequence and outputs their location and category~\cite{bernardin2008evaluating}. 
%
%
%
%
While prior work~\cite{segu2023darth} explored domain adaptation of \mot to diverse conditions, continual learning for \mot would provide a flexible and inexpensive solution to incrementally expand the \ac{mot} model to new classes according to the changing necessities.
For example, as illustrated in Fig.~\ref{fig:teaser}, one can train an MOT model to track cars and then expand its functionality to track pedestrians with new training data only annotated for pedestrians.

Following the tracking-by-detection paradigm~\cite{ramanan2003finding}, most MOT systems first detect object locations and classes via an object detector, and then associate the detected instances across frames via a data association module.
State-the-art trackers often use a combination of motion and appearance cues in their association module~\cite{aharon2022bot,wang2022smiletrack,zhang2022bytetrack}.
While motion cues are straight-forward to use with simple heuristics, appearance cues are used for object re-identification (Re-ID) and are more robust to complex object motion and large object displacement across adjacent frames. 
Appearance-based association typically requires a Re-ID module~\cite{du2023strongsort,pang2021quasi,Wojke2017simple} for learning Re-ID features. 
However, it is crucial to make such learned appearance representations flexible to incrementally added categories.
Training the appearance extractors only on the new classes would indeed results in catastrophic forgetting of Re-ID features for older classes, and degrade the association performance (Tab. \ref{tab:bdd_exp}, Fine-tuning).
Although previous work~\cite{peng2020faster,shmelkov2017incremental,zhou2020lifelong} explores class-incremental learning of object detectors, these approaches are sub-optimal for MOT by not addressing the data association stage.
%
%

To address this problem, we introduce COOLer, a COntrastive- and cOntinual-Learning-based multiple object tracker.
Building on the state-of-the-art appearance-based tracker QDTrack~\cite{pang2021quasi}, COOLer represents the first comprehensive approach for continual learning for appearance-based trackers by addressing class-incremental learning of both the building blocks of an MOT system, \emph{i.e.} object detection and data association.
To continually learn to track new categories while preventing catastrophic forgetting, we propose to combine the available ground truth labels from the newly added categories with the association pseudo-labels and the temporally-refined detection pseudo-labels generated by the previous-stage tracker on the new training data. 
%
%
Furthermore, adding classes incrementally without imposing any constraint may cause overlapping instance representations from different classes, blurring the decision boundaries and leading to misclassifications. 
While traditional contrastive learning can disentangle the representations of different classes, they undermine the intra-class discrimination properties of the instance embeddings for data association.
%
To this end, we propose a novel contrastive class-incremental instance representation learning formulation that pushes the embedding distributions of different classes away from each other while keeping the embedding distributions of the same class close to a Gaussian prior. 
%
To assess the effectiveness of continual learning strategies for \mot, we propose a practical and comprehensive evaluation protocol and conduct extensive experiments on the \bdd~\cite{yu2020bdd100k} and SHIFT~\cite{shift2022} datasets. 
%

We demonstrate that COOLer can alleviate forgetting of both tracking and detection, while effectively acquiring incremental knowledge.
Our key contributions are: (i) we introduce COOLer, the first comprehensive method for class-incremental learning for multiple object tracking; (ii) we propose to use the previous-stage tracker to generate data association pseudo-labels to address catastrophic forgetting of association of previous classes and leverage the temporal information to refine detection pseudo-labels; (iii) we introduce class-incremental instance representation learning to disentangle class representations and further improve both detection and association performance.

\section{Related Work}
Continual learning aims at learning new knowledge continually while alleviating forgetting. 
Various continual learning strategies have been proposed, including model growing~\cite{rusu2016progressive}, regularization~\cite{kirkpatrick2017overcoming,li2017learning}, parameter isolation~\cite{mallya2018packnet}, and replay~\cite{rebuffi2017icarl}. 
%
We here discuss related literature in continual learning for object detection, unsupervised Re-ID learning, and contrastive representation learning.

\noindent\textbf{Continual Learning for Object Detection.} 
As opposed to MOT, object detection under limited labels~\cite{shmelkov2017incremental,liu2023continual,peng2020faster} and diverse conditions~\cite{fan2022towards,fan2022normalization} is widely studied.
Shmelkov et al.~\cite{shmelkov2017incremental} propose the first method for continual learning for object detection. 
It uses the old model as the teacher model which generates pseudo labels for the classification and bounding box regression outputs to prevent forgetting. 
Later works~\cite{liu2023continual,peng2020faster} follow this diagram by incorporating the state-of-the-art detectors such as Faster R-CNN~\cite{ren2015faster} and Deformable DETR~\cite{zhu2020deformable}. 
While our work also uses detection pseudo-labels, we refine them temporally by leveraging a multiple-object tracker.

\begin{figure*}[t]
\centering
\includegraphics[width=1.0\textwidth]{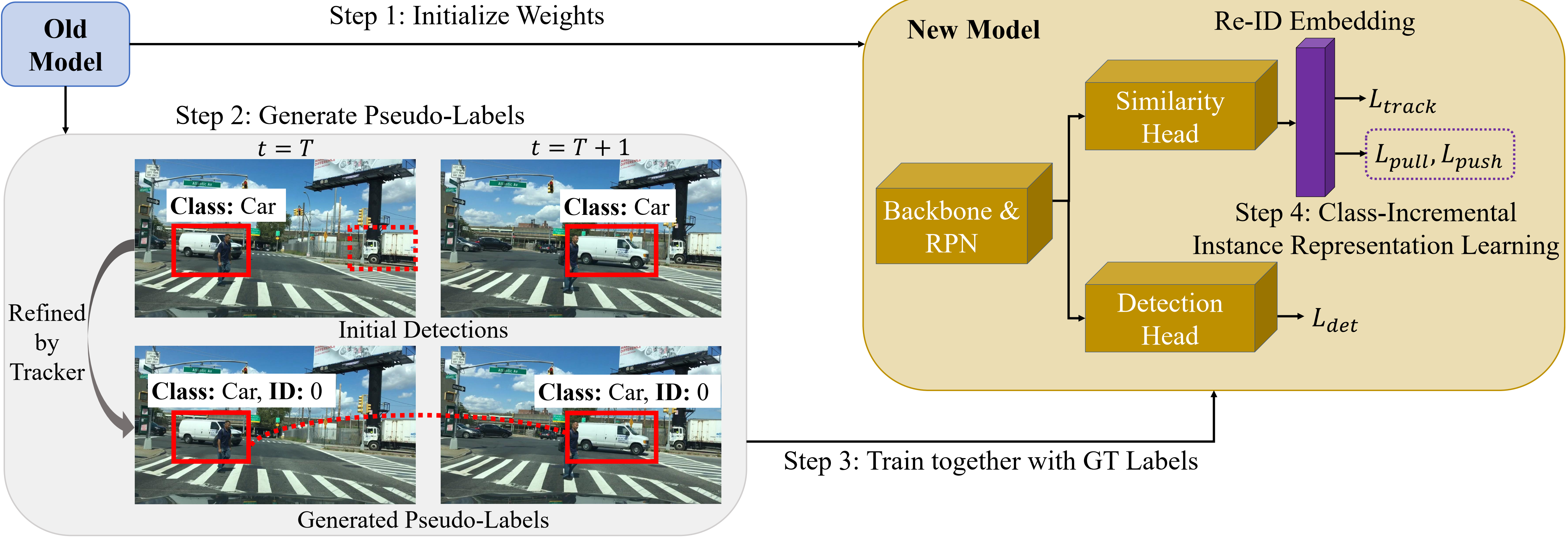} 
\caption{Pipeline of COOLer for class-incremental learning. 1). Initialize the new model with weights from the old model. 2). Use the previous tracker to refine the initial detections and generate detection and association pseudo-labels. 3). Append them to the ground truth labels to train the new model jointly. 4). During training, apply class-incremental instance representation learning.} 
\label{fig:method}
\end{figure*}

\noindent\textbf{Unsupervised Re-ID Learning.} As annotating instance IDs is laborious and time-consuming, unsupervised Re-ID learning proposes to learn data association from video sequences without annotations given only a pre-trained detector~\cite{karthik2020simple}. Most unsupervised Re-ID learning approaches generate pseudo-identities to train the association module from a simple motion-based tracker~\cite{karthik2020simple}, image clustering
~\cite{fan2018unsupervised,li2019unsupervised,wu2020tracklet} or contrastive learning of instance representation under data augmentation~\cite{segu2023darth}. 
In contrast, our class-incremental instance representation learning approach handles a combination of labeled and unlabeled Re-ID data to continually learn Re-ID of new categories' instances without forgetting the old ones.

\noindent\textbf{Contrastive Representation Learning.} 
Contrastive learning~\cite{chen2020simple,he2020momentum} aims to attract representations of similar samples and push away representations of dissimilar ones. 
Previous continual learning methods leverage contrastive learning at a category level. Mai et al.~\cite{mai2021supervised} propose supervised contrastive replay and use a nearest-class-mean classifier instead of the softmax classifier. 
Co2L~\cite{cha2021co2l} shows that contrastively-learned representations are more robust to catastrophic forgetting than ones trained with cross-entropy.
OWOD~\cite{joseph2021towards} introduces a memory queue for updating the class mean prototype vector during training to help contrastive learning. 
However, such class-contrastive formulations often collapse intra-class representations, hindering Re-ID-based data association in MOT. Our class-incremental instance representation learning approach maintains intra-class variability with a contrastive loss that estimates standard deviation prototype vectors for each class and keeps the class distribution close to a prior.  


\section{Method}
We define the continual learning problem for MOT (Sec.~\ref{sec:overview}). We then provide an overview of COOLer (Sec.~\ref{sec:overview}) and introduce its two key components, namely continual pseudo-label generation for detection and data association (Sec.~\ref{sec:track_pl}), and class-incremental instance representation learning (Sec.~\ref{sec:contrast}).

\subsection{Problem Definition}
\label{sec:problem}
We define continual learning for MOT as a class-incremental learning (CIL) problem over a sequence of B training stages $\{\mathcal{S}^0, \mathcal{S}^1, ..., \mathcal{S}^{B-1}\}$, where at each stage $b$ a set of categories $Y_b$ is introduced. 
$\mathcal{S}^b = \{\mathcal{X}^b, \mathcal{D}^b, \mathcal{T}^b\}$ is the set of training videos $\mathcal{X}^b$, detection labels $\mathcal{D}^b$, and tracking labels $\mathcal{T}^b$ for a set of categories $Y_b$ at stage $b$.
Although typical CIL assumes no overlapping classes in different tasks $b$ and $b'$, it is common in real-world applications to observe old classes in new stages~\cite{xie2022general}.
Thus, we assume that categories from another stage $b'$ may occur again at $b$ despite not being in the annotation set, i.e. $Y_b \cap Y_{b'} = \emptyset$. 
The goal is continually learning an MOT model that can track $Y_b$ without forgetting to track $\bar{Y}_{b-1} = Y_0 \cup ... \cup {Y}_{b-1}$.  
During each stage $b$, only data $\mathcal{S}^b$ can be accessed.
After each training stage $b$, the model is evaluated over all seen classes $\mathcal{Y}_b = Y_0 \cup ... \cup Y_b$.

\subsection{COOLer}
\label{sec:overview}
%
\textbf{Architecture.} COOLer's architecture is based on the representative appearance-based tracker QDTrack~\cite{pang2021quasi}, which consists of a Faster R-CNN~\cite{ren2015faster} object detector and a similarity head to learn Re-ID embeddings for data association.

\noindent\textbf{Base Training.} Given the data $\mathcal{S}^0 = \{\mathcal{X}^0, \mathcal{D}^0, \mathcal{T}^0\}$ from the first stage $b=0$, we train the base model $\phi^{0}$ following QDTrack. Let $\mathcal{\hat{D}}^0$ be the detector predictions and $\mathcal{\hat{V}}^0$ their corresponding Re-ID embeddings.
QDTrack is optimized end-to-end with a detection loss $\mathcal{L}_{\mathrm{det}}$ to train the object detector, and a tracking loss $\mathcal{L}_{\mathrm{track}}$ to learn the Re-ID embeddings for data association. $\mathcal{L}_{\mathrm{det}}$ is computed from $\mathcal{D}^0$ and  $\mathcal{\hat{D}}^0$ as in Faster R-CNN~\cite{ren2015faster}. As for the tracking loss $\mathcal{L}_{\mathrm{track}}$, QDTrack first samples positive and negative pairs of object proposals in adjacent frames using $\mathcal{D}^0$, $\mathcal{T}^0$, and $\mathcal{\hat{D}}^0$. Then, $\mathcal{L}_{\mathrm{track}}$ is computed from a contrastive loss using the Re-ID embeddings $\mathcal{\hat{V}}^0$ of the sampled proposals to cluster object embeddings of the same IDs and separate embeddings of different instances. Refer to the original QDTrack paper~\cite{pang2021quasi} for more details. The final loss is:
\begin{equation}
    \mathcal{L}^{0} = \mathcal{L}_{\mathrm{det}}(\mathcal{\hat{D}}^0, \mathcal{D}^0) + \mathcal{L}_{\mathrm{track}}(\mathcal{\hat{D}}^0,   \mathcal{\hat{V}}^0, \mathcal{D}^0,
\mathcal{T}^0) .
\end{equation}

\noindent\textbf{Continual Training.} Given the old model $\phi^{b-1}$ trained up to the stage $b-1$, and the new data $\mathcal{S}^b$ for the stage $b$, COOLer is the first tracker to incrementally learn to track the new classes $Y_b$ without forgetting the old ones $\bar{Y}_{b-1}$.
%
We propose a continual pseudo-label generation strategy for MOT (Sec.~\ref{sec:track_pl}) that uses the previous tracker $\phi^{b-1}$ to generate pseudo-labels $\{\bar{\mathcal{D}}^b_{\mathrm{old}}, \bar{\mathcal{T}}^b_{\mathrm{old}}\}$ for the old classes $\bar{Y}_{b-1}$, and combine them with the ground-truth labels $\{\mathcal{D}^b_{\mathrm{new}}, \mathcal{T}^b_{\mathrm{new}}\}$ for the new ones ${Y}_{b}$ to train the new model $\phi^{b}$.   
%
%
To further disentangle the Re-ID embedding space for different classes and instances, we propose a novel class-incremental instance representation learning approach (Sec.~\ref{sec:contrast}).
See Fig.~\ref{fig:method} for an overview.  
 
\subsection{Continual Pseudo-label Generation for Tracking}
\label{sec:track_pl}
While training with detection pseudo-labels generated by the previous object detector has proven effective against catastrophic forgetting in CIL of object detection~\cite{peng2020faster,zheng2021contrast}, detection pseudo-labels lack the instance association information, which is crucial to learn the Re-ID module in appearance-based MOT. 
%
%
We instead propose to use the MOT model $\phi^{b-1}$ from the previous stage ${b-1}$ to simultaneously generate temporally-refined detection pseudo-labels $\bar{\mathcal{D}}^b_{\mathrm{old}}$ and instance association pseudo-labels $\bar{\mathcal{T}}^b_{\mathrm{old}}$ for the old classes $\bar{Y}_{b-1}$ in the new stage $b$.
We then train the new tracker $\phi^{b}$ on the union of the pseudo-labels $\{\bar{\mathcal{D}}^b_{\mathrm{old}}, \bar{\mathcal{T}}^b_{\mathrm{old}}\}$  for old classes $\bar{Y}_{b-1}$ and ground-truth labels $\{\mathcal{D}^b_{\mathrm{new}}, \mathcal{T}^b_{\mathrm{new}}\}$ for new classes ${Y}_{b}$:
\begin{equation}
\label{eqn:pl}
     \mathcal{L}^b_{\mathrm{pseudo}} = \mathcal{L}_{\mathrm{det}}(\mathcal{\hat{D}}^b, \bar{\mathcal{D}}^b_{\mathrm{old}} \cup \mathcal{D}^b_{\mathrm{new}}) + \mathcal{L}_{\mathrm{track}}(\mathcal{\hat{D}}^b, \mathcal{\hat{V}}^b,  \bar{\mathcal{D}}^b_{\mathrm{old}} \cup \mathcal{D}^b_{\mathrm{new}}, \bar{\mathcal{T}}^b_{\mathrm{old}} \cup \mathcal{T}^b_{\mathrm{new}}) .
\end{equation}
It is worth noticing that, unlike detection pseudo-labels in~\cite{peng2020faster,zheng2021contrast}, our detection pseudo-labels are temporally refined by the tracking algorithm, resulting in a reduced number of false positives and in recovery of initially missed detections.
Moreover, the pseudo-identities $\bar{\mathcal{T}}^b_{\mathrm{old}}$ alleviate catastrophic forgetting in data association by training the similarity head on old classes $\bar{Y}_{b-1}$.

\begin{figure*}[t]
\centering
\includegraphics[width=1.0\textwidth]{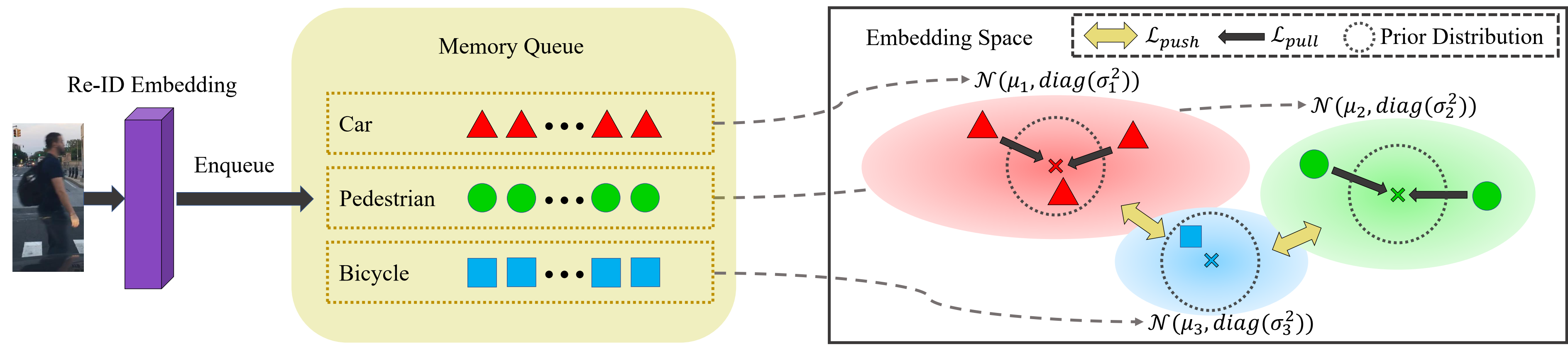} 
\caption{Illustration of our class-incremental instance representation learning. We keep a memory queue to update the class embedding distributions. The contrastive loss includes an inter-class pushing loss and an intra-class pulling loss. } 
\label{fig:contrast_loss}
\end{figure*}
\subsection{Class-Incremental Instance Representation Learning}
\label{sec:contrast}
%
Our class-incremental learning strategy based on tracking pseudo-labels (Sec. \ref{sec:track_pl}) enforces that each instance must be well-separated from others in the embeddings space, but does not constrain where Re-ID features for each class are projected, potentially leading to entangled class distributions that hurt both detection and tracking performance.
Previous CIL approaches~\cite{cha2021co2l,joseph2021towards,mai2021supervised} ensure separation of class distributions by applying class-contrastive losses during incremental learning.
However, naively applying contrastive learning on the instance embedding space would cause the distribution of a class' embeddings to collapse to a single point, undermining the intra-class discrimination properties of the learned Re-ID embeddings necessary for effective data association.

To this end, we introduce a novel contrastive loss for class-incremental instance representation learning that disentangles embeddings of different classes while maintaining the intra-class variability of the embeddings (Fig.~\ref{fig:contrast_loss}). 
%

\noindent\textbf{Class Prototype Vectors.} First, we model variability of instance embeddings within each class $c$ by approximating each class' embedding distribution as a Gaussian $\mathcal{N}(\boldsymbol{\mu}_c, \mathrm{diag}(\boldsymbol{\sigma}_c^2))$, whose class mean prototype vector $\boldsymbol{\mu}_c$ and class standard deviation prototype vector $\boldsymbol{\sigma}_c$ are approximated online as the exponential moving average of a memory queue with limited size $N_{\mathrm{queue}}$ that stores exemplary class embeddings.
See Supplement Sec. A for details on the memory queue.
%
%
%

%

\noindent\textbf{Contrastive Loss.} Our contrastive loss 
consists of a pushing term $\mathcal{L}_{\mathrm{push}}$ that pushes distributions of different classes away from each other, and a pulling term $\mathcal{L}_{\mathrm{pull}}$ that keeps the class distribution close to a prior, ensuring intra-class variability. 
We derive such losses from the Bhattacharyya distance $D_{B}$, which measures the similarity between distributions $\mathcal{N}(\boldsymbol{\mu}_{c_1}, \mathrm{diag}(\boldsymbol{\sigma}_{c_1}^2))$ and $\mathcal{N}(\boldsymbol{\mu}_{c_2}, \mathrm{diag}(\boldsymbol{\sigma}_{c_2}^2))$ of two classes $c_1$ and $c_2$. Their Bhattacharyya distance is:
\begin{align}
\label{eq:db}
\small
\begin{split}
    D_{B}(c_1,c_2) = &\frac{1}{8} (\boldsymbol{\mu}_{c_1}-\boldsymbol{\mu}_{c_2})^{T} \boldsymbol{\Sigma}^{-1}_{c_1,c_2} (\boldsymbol{\mu}_{c_1}-\boldsymbol{\mu}_{c_2}) + \frac{1}{2} \ln{\frac{\det \boldsymbol{\Sigma}_{c_1,c_2}}{\sqrt{\det\boldsymbol{\Sigma}_{c_1} \det\boldsymbol{\Sigma}_{c_2}}}},
\end{split}
\end{align}
where $\boldsymbol{\Sigma}_{c_1,c_2} = \frac{\boldsymbol{\Sigma}_{c_1} + \boldsymbol{\Sigma}_{c_2}}{2}$.
As it is hard to back-propagate gradients for the prototype mean and standard deviation $\boldsymbol{\mu}_{c}, \boldsymbol{\Sigma}_{c}$, we additionally introduce the per-batch embedding mean $\bar{\boldsymbol{\mu}}_c$ and standard deviation $\bar{\boldsymbol{\sigma}}_c$ for class $c$:
\begin{equation}
\small
    \bar{\boldsymbol{\mu}}_c = \frac{1}{N_c} \sum_{i=1}^{N_c} \boldsymbol{v}_{c,i},
     \bar{\boldsymbol{\sigma}}_c = \sqrt{\frac{1}{N_c} \sum_{i=1}^{N_c} (\boldsymbol{v}_{c,i}-\boldsymbol{\mu}_c)^2},
\end{equation} 
where $N_c$ denotes the number of embedding vectors for class $c$ in the current batch and $\boldsymbol{v}_{c,i}$ is the $i_\mathrm{th}$ embedding vector for class $c$.
 
%

\noindent\textbf{Pushing Loss.} For the pushing loss, the distance between the two class distributions is derived from the first term of Eqn.~\ref{eq:db} as:
\begin{equation}
\small
D_{\mathrm{push}}(c_1, c_2) = \sqrt{(\bar{\boldsymbol{\mu}}_{c_1}-\boldsymbol{\mu}_{c_2})^{T} \boldsymbol{\Sigma}^{-1}_{c_1,c_2} (\bar{\boldsymbol{\mu}}_{c_1}-\boldsymbol{\mu}_{c_2}) }. 
\end{equation}
We use the following hinge-based pushing loss to separate the two distributions:
\begin{equation}
\small
\mathcal{L}_{\mathrm{push}} = \frac{1}{C(C-1)} \sum_{c_1=1}^{C} \sum_{\substack{c_2=1 \\ c_2\neq c_1}}^C  [\Delta_{\mathrm{push}} - D_{\mathrm{push}}(c_1,c_2)]_+^2,
\end{equation}
where $C$ is the number of classes, $\Delta_\mathrm{push}$ is the hinge factor, and $[x]_+ = \mathrm{max}(0, x)$. 

%

\noindent\textbf{Pulling Loss.} For the pulling loss, we introduce a prior Gaussian distribution $\mathcal{N}(\boldsymbol{\mu}_c, \mathrm{diag}(\boldsymbol{\sigma}_p^2))$ for each class, which has the same class mean prototype vector $\boldsymbol{\mu}_c$ while the standard deviation $\boldsymbol{\sigma}_p$ is fixed. 
%
We derive the distance between the class distribution and the prior distribution from the second term of Eqn.~\ref{eq:db} as:
\begin{equation}
\label{eq:db_pull}
\small
D_\mathrm{pull}(c,p) = \frac{1}{2}( \sum_{j=1}^{N_d} \ln (
  \frac{\bar{\sigma}_{c,j}^2 + \sigma_{p,j}^2 } {2}) - \sum_{j=1}^{N_d} \ln (
  \bar{\sigma}_{c,j}\sigma_{p,j})),
\end{equation}
where $N_d$ is the dimension of the embedding.
We find that directly applying Eqn.~\ref{eq:db_pull} as the pulling loss will lead to numerical instability during optimization, as the logarithm operator is non-convex. 
We propose the following surrogate based on the $\mathcal{L}_2$ distance for a smoother optimization landscape as follows:
\begin{equation}
\small
    \mathcal{L}_{\mathrm{pull}}  = \frac{1}{C} \sum_{c=1}^{C} \sum_{j=1}^{N_d} ( \bar{\sigma}_{c,j} - \sigma_{p,j})^2.
\end{equation}
%
%
%

\noindent\textbf{Total Loss.} Finally, we extend Eqn.~\ref{eqn:pl} with our pulling and pushing contrastive losses to learn the tracking model $\phi^b$ at stage $b$:
\begin{equation}
\small
    \mathcal{L}^b = \mathcal{L}_{\mathrm{det}} + \mathcal{L}_{\mathrm{track}} + \beta_1 \mathcal{L}_{\mathrm{pull}}  + \beta_2 \mathcal{L}_{\mathrm{push}},
\end{equation}
where $\beta_1$ and $\beta_2$ are weights for the pushing and the pulling loss respectively.
%

%

%

%
%

\section{Evaluation Protocol}
We introduce a protocol for evaluating algorithms for class-incremental MOT.

\noindent\textbf{Datasets.} We use the \bdd~\cite{yu2020bdd100k} and \shift~\cite{shift2022} tracking datasets for evaluation. 
\bdd is a large-scale real-world driving dataset for MOT with 8 classes.
\shift is a large-scale synthetic driving dataset for MOT with 6 classes. 
Because of the size of the SHIFT dataset, training multiple stages on it is not feasible with modest computational resources. To ensure practicality for all researchers', we propose to only use its clear-daytime subset.
The detailed class statistics for each dataset are reported in Tab.~\ref{tab:statistics}.
Note that other popular MOT datasets are unsuitable for our setting. The MOT20 dataset~\cite{dendorfer2020mot20} has very few categories. While TAO~\cite{dave2020tao} has
hundreds of classes, due to the scarcity of annotations it is intended as an evaluation benchmark and is not suitable
for CIL. 

\begin{table}[htbp]
\centering
\scriptsize
\setlength\tabcolsep{2pt}
\caption{Class frequencies for the training splits of BDD100K~\cite{yu2020bdd100k} and SHIFT's~\cite{shift2022} clear-daytime subset. }
\label{tab:statistics}
\renewcommand\arraystretch{0.8}
\begin{tabular}{@{}ccccccccc@{}}
\toprule
 Dataset              & Car     & Ped & Truck  & Bus   & Bike  & Rider & Motor & Train \\ \midrule
BDD100K~\cite{yu2020bdd100k}        & 2098517 & 369798     & 149411 & 57860 & 25565 & 20107 & 12176 & 1620  \\
SHIFT~\cite{shift2022} & 677580  & 749640     & 145940 & 65367 & 52267 & -     & 74469 & -     \\ \bottomrule
\end{tabular}

\end{table}

%

\noindent\textbf{Protocol.} The choice of the class ordering during incremental stages may largely impact results and observations.
Object detection benchmarks typically add classes by their alphabetical order.
However, in real-world MOT applications the annotation order often depends by (i) class frequency or (ii) semantic grouping.
%
We hence propose the practical class splits to mirror the practitioners' needs. 
First, we propose two \textit{frequency-based splits}. The Most$\rightarrow$Least (\textbf{M}$\rightarrow$\textbf{L}) split incrementally adds classes one-by-one from the most to the least frequent class according to Tab.~\ref{tab:statistics}. 
General$\rightarrow$Specific (\textbf{G}$\rightarrow$\textbf{S}) only evaluates one incremental step by dividing the classes into two groups: the first half of the most populated classes (General) and the remainder (Specific).
%
Then, we propose a \textit{semantic split}.
We group classes into three super-categories according to their semantic similarity: vehicles, bikes, and humans. 
Therefore, we experiment on the Vehicle$\rightarrow$Bike$\rightarrow$Human (\textbf{V}$\rightarrow$\textbf{B}$\rightarrow$\textbf{H}) setting with two incremental steps. 

Taking \bdd as example, in the \textbf{M}$\rightarrow$\textbf{L} setting the classes are added as follows:
car$\shortrightarrow$pedestrian$\shortrightarrow$truck$\shortrightarrow$bus$\shortrightarrow$bike$\shortrightarrow$rider$\shortrightarrow$motor$\shortrightarrow$train. 
In the \textbf{G}$\rightarrow$\textbf{S}  setting, the model is first trained on \{car, pedestrian, truck, bus\}, and then \{bike, rider, motor, train\} are added at once. In the \textbf{V}$\rightarrow$\textbf{B}$\rightarrow$\textbf{H} setting, the classes are added as follows: \{car, truck, bus, train\}$\shortrightarrow$\{bike, motor\}$\shortrightarrow$\{pedestrian, rider\}.

\section{Experiments}

\subsection{Baselines}
Since no prior work studied class-incremental learning for multiple objcet tracking, we compare COOLer with the following baseline methods:

\noindent\textbf{Fine-tuning}. In each incremental step, the model is trained only on the training data of the new classes, without addressing catastrophic forgetting.
\begin{table*}[h!]
\centering
\scriptsize
\setlength\tabcolsep{2pt}
\renewcommand\arraystretch{0.6}
\caption{ \textbf{Class-incremental Learning on \bdd.} We conduct experiments on \textbf{M}$\rightarrow$\textbf{L}, \textbf{G}$\rightarrow$\textbf{S} and \textbf{V}$\rightarrow$\textbf{B}$\rightarrow$\textbf{H} settings.
We compare COOLer with the Fine-tuning, Distillation, Det PL baselines and the oracle tracker.}
\label{tab:bdd_exp}
\begin{tabular}{@{}llccccccc@{}}
\toprule \toprule
\multirow{2}{*}{\begin{tabular}[c]{@{}l@{}}\textbf{Setting}\\ Stage (+New Classes)\end{tabular}}                                              &       & \multicolumn{7}{c}{All Classes}                                                                                         \\ \cmidrule(l){3-9} 
&Method                                                                   & mMOTA          & mHOTA          & mIDF1   & MOTA & HOTA & IDF1        & mAP                    \\ \midrule \midrule

\textbf{M}$\rightarrow$\textbf{L}  Stage 0 (Car) &   &  67.6 & 62.1 & 73.3 &  67.6 & 62.1 & 73.3 & 58.7 \\
\midrule
\multirow{5}{*}{\begin{tabular}[c]{@{}l@{}}\textbf{M}$\rightarrow$\textbf{L}\\ Stage 1 (+Pedestrian) \end{tabular}}     & Fine-tuning    & 15.6      & 21.8       &27.7    & 4.5 & 19.1 & 14.1&      19.9                  \\ 
                                                                                                & Distillation &   46.4        &  51.7        & 63.0       & 61.8 & 59.0 & 70.1  & 47.1                      \\  
                                                                                                & Det PL      &   46.7       &  49.5        & 59.2  &56.3& 53.9&  61.8       &  46.8                           \\  
                                                                                                & COOLer        & \textbf{54.2}  &  \textbf{52.6}& \textbf{64.3} &\textbf{62.7}&\textbf{59.5}&\textbf{70.5}& \textbf{47.4}  \\  
                                                                                               &  \cellcolor{lightgray}Oracle       & \cellcolor{lightgray}57.4 & \cellcolor{lightgray}53.6 & \cellcolor{lightgray}65.9 & \cellcolor{lightgray}65.1 &\cellcolor{lightgray}59.9&                                                                          \cellcolor{lightgray}71.5   & \cellcolor{lightgray}48.3            \\\midrule 
\multirow{5}{*}{\begin{tabular}[c]{@{}l@{}}\textbf{M}$\rightarrow$\textbf{L}\\ Stage 2 (+Truck)\end{tabular}}          & Fine-tuning    &   -11.5        & 12.8         & 14.0  & -2.2 & 13.3 & 6.2         &  11.7           \\ 
                                                                                                & Distillation &     27.3     &   47.8     &    57.3   &56.9 & 56.6 & 67.4    &  \textbf{42.8}                      \\  
                                                                                                & Det PL      &  34.9         & 47.1          &55.6         &57.3 & 55.5  &  65.1 & 42.5              \\  
                                                                                                & COOLer      & \textbf{42.8}  & \textbf{49.2} &  \textbf{59.6}     &  \textbf{58.6}&\textbf{57.9}&\textbf{68.7}&42.6           \\ 
                                                                                                 & \cellcolor{lightgray}Oracle      & \cellcolor{lightgray}49.8 & \cellcolor{lightgray}50.8 &  \cellcolor{lightgray}62.1     &  \cellcolor{lightgray}63.2 &\cellcolor{lightgray}58.9&\cellcolor{lightgray}70.4&\cellcolor{lightgray}45.0           \\ \midrule
\multirow{5}{*}{\begin{tabular}[c]{@{}l@{}}\textbf{M}$\rightarrow$\textbf{L}\\ Stage 3  (+Bus)\end{tabular}}            & Fine-tuning           &  -24.0           & 9.2        & 9.3   & -2.0 & 8.6 & 2.5  &  9.9           \\  
                                                                                                & Distillation           &  -11.2         &   43.4         &  50.8   & 54.0 & 55.1 & 65.6       &  40.8                \\  
                                                                                                & Det PL       &  14.1         &    43.1     &     49.1      & 53.6 &  53.4&  61.5 &   40.9         \\  
                                                                                                & COOLer        &  \textbf{34.0} &  \textbf{47.9} & \textbf{57.3} &\textbf{55.8}&\textbf{56.8}& \textbf{67.4}& \textbf{41.9}     \\
                                                                                                & \cellcolor{lightgray}Oracle      & \cellcolor{lightgray}45.4  & \cellcolor{lightgray}50.1  &  \cellcolor{lightgray}60.9 &\cellcolor{lightgray}62.5&\cellcolor{lightgray}58.7&\cellcolor{lightgray}70.2&\cellcolor{lightgray}44.5       \\\midrule
\multirow{5}{*}{\begin{tabular}[c]{@{}l@{}}\textbf{M}$\rightarrow$\textbf{L}\\ Stage 4  (+Bicycle)\end{tabular}}        & Fine-tuning          &  -16.0           & 5.6          & 6.5   & -0.8 &  4.2 & 0.8      &4.6            \\ 
                                                                                                & Distillation          &    -19.4        &   40.1         &   47.1 & 51.9 & 53.2 & 63.6         &  35.1                               \\ 
                                                                                                & Det PL           &   3.6       &   37.6        &  42.4       &  41.2  & 43.0 &  46.2&  34.3           \\  
                                                                                                & COOLer       & \textbf{28.6}   &   \textbf{44.4}&  \textbf{53.9} &\textbf{53.2}  & \textbf{55.7} & \textbf{66.1}  &  \textbf{36.5}           \\
                                                                                                   & \cellcolor{lightgray}Oracle       & \cellcolor{lightgray}41.3  & \cellcolor{lightgray}47.1  & \cellcolor{lightgray}58.0  & \cellcolor{lightgray}62.2 &   \cellcolor{lightgray}58.5                                                                          &   \cellcolor{lightgray}69.9                                                &\cellcolor{lightgray}39.9       \\\midrule \midrule
 \textbf{G}$\rightarrow$\textbf{S}  Stage 0 (General)  & &   45.6  &  50.3 &  61.1 & 62.4& 59.0&70.2&  44.6  \\
\midrule                                                                                                 
\multirow{5}{*}{\begin{tabular}[c]{@{}l@{}}\textbf{G}$\rightarrow$\textbf{S}\\ Stage 1 (+Specific)\end{tabular}} & Fine-tuning              &   -24.7      & 11.7         & 14.1       &-0.5&5.7& 1.5  &  8.4               \\  
                                                                                                & Distillation          &    -32.9      &  35.4         &  42.4       &59.6 &57.3& 68.3 & 29.5         \\  
                                                                                                & Det PL                &   6.0        &  34.9         & 41.5      &54.1&52.0& 59.6  &  27.9       \\  
                                                                                                & COOLer         & \textbf{28.6} &  \textbf{38.5}&  \textbf{48.2} &\textbf{60.5}&\textbf{58.2}&\textbf{69.3}&  \textbf{29.9}              \\ 
                                                                                                & \cellcolor{lightgray}Oracle         &  \cellcolor{lightgray}30.4 & \cellcolor{lightgray}38.9  &  \cellcolor{lightgray}49.0      &\cellcolor{lightgray}61.8  &\cellcolor{lightgray}58.7  &\cellcolor{lightgray}70.0  &\cellcolor{lightgray}30.9    \\ \midrule \midrule
                                                                                               \textbf{V}$\rightarrow$\textbf{B}$\rightarrow$\textbf{H}  Stage 0 (Vehicle) & & 33.1    & 39.2  & 46.8 & 65.1&61.0& 72.3 & 35.0   \\
\midrule
\multirow{5}{*}{\begin{tabular}[c]{@{}l@{}}\textbf{V}$\rightarrow$\textbf{B}$\rightarrow$\textbf{H}\\ Stage 1 (+Bike)\end{tabular}}   & Fine-tuning            &   -27.5     &  9.7          & 11.2        &-0.6&4.9& 1.2   &    7.3             \\ 
                                                                                                & Distillation          &  -30.5         & 34.1          &  39.6 &62.9&59.7&70.6& \textbf{29.1}              \\  
                                                                                                & Det PL               &      -3.3     &     31.3      &    37.0     &53.7&51.4& 57.9 &    26.7      \\  
                                                                                                & COOLer        & \textbf{24.4}  & \textbf{36.8} &   \textbf{44.8}        &\textbf{63.1}&\textbf{60.2}&\textbf{70.9}&    \textbf{29.1}             \\ 
                                                                                                & \cellcolor{lightgray}Oracle        & \cellcolor{lightgray}27.6 & \cellcolor{lightgray}38.5 & \cellcolor{lightgray}47.8    & \cellcolor{lightgray}64.3 &\cellcolor{lightgray}60.4 &\cellcolor{lightgray}71.6                                                                                &\cellcolor{lightgray}30.8              \\ \midrule
\multirow{5}{*}{\begin{tabular}[c]{@{}l@{}}\textbf{V}$\rightarrow$\textbf{B}$\rightarrow$\textbf{H}\\ Stage 2 (+Human)\end{tabular}}  & Fine-tuning        &  7.4        & 10.0          & 13.1       &4.4&18.2&  13.4  &  8.2      \\  
                                                                                                & Distillation         &  14.8         &   35.8        & 43.9      & 55.9&55.9& 66.4   &   27.9               \\  
                                                                                                & Det PL         & 15.5          &   34.6        &   41.5      &52.1&51.4& 58.7 &   27.4       \\  
                                                                                                & COOLer       & \textbf{27.1} &  \textbf{37.5}&  \textbf{46.8}& \textbf{59.2}&\textbf{57.8}&\textbf{68.8}& \textbf{28.8}           \\ 
                                                                                                 &\cellcolor{lightgray}Oracle       & \cellcolor{lightgray}30.4 & \cellcolor{lightgray}38.9 &  \cellcolor{lightgray}49.0 &\cellcolor{lightgray}61.8  &\cellcolor{lightgray}58.7 &\cellcolor{lightgray}70.0 &\cellcolor{lightgray}30.9       \\ \bottomrule \bottomrule 
\end{tabular}

\end{table*}

\noindent\textbf{Distillation}. We design a distillation baseline based on Faster-ILOD~\cite{peng2020faster}, a state-of-the-art class-incremental object detection method that uses distillation losses from a teacher model of the previous stage to alleviate forgetting.
To further address forgetting in data association, we add the following distillation loss on the similarity head of QDTrack to enforce the cosine similarity between teacher and student embeddings for the old classes:
\begin{equation}
\small
        \mathcal{L}_{\text{track}}^{\text{dist}} =  
    (\frac{\mathbf{v}_{\text{teacher}} \cdot \mathbf{v}_{\text{student}}}{\norm{\mathbf{v}_{\text{teacher}}}_2 \cdot \norm{\mathbf{v}_{\text{student}}}_2}-1)^2,
\end{equation}
where $\mathbf{v}_{\text{teacher}}$ and $\mathbf{v}_{\text{student}}$ are the Re-ID embeddings of the teacher and the student model, computed from the same proposals sampled for Faster-ILOD's ROI head distillation. 
%
The final loss is then ${\mathcal{L} = \mathcal{L}_{\mathrm{det}} + \mathcal{L}_{\mathrm{track}}+ \mu_1 \mathcal{L}_{\text{det}}^{\text{dist}} +  \mu_2 \mathcal{L}_{\text{track}}^{\text{dist}}}$, 
where $\mathcal{L}_{\text{det}}^{\text{dist}}$ is the detection distillation loss in~\cite{peng2020faster}, and $\mu_1,\mu_2$ are set to 1.

\noindent\textbf{Detection Pseudo-Labels (Det PL)}. We compare against a baseline that only trains on the joint set of ground-truth labels for the new classes and high-confident ($>0.7$) detection pseudo-labels from the old detector for the old classes. Unlike our method, this baseline does not temporally refine the detection pseudo-labels with the tracker, and does not provide association pseudo-labels.

\noindent\textbf{Oracle}. We compare the result with an oracle tracker trained in a single stage on the ground truth annotations of all classes.

\begin{table*}[htbp]
\centering
\scriptsize
\setlength\tabcolsep{2pt}
\renewcommand\arraystretch{0.6}
\caption{\textbf{Class-incremental Learning on \shift.} We conduct experiments on \textbf{M}$\rightarrow$\textbf{L}, \textbf{G}$\rightarrow$\textbf{S} and \textbf{V}$\rightarrow$\textbf{B}$\rightarrow$\textbf{H} settings.
We compare COOLer with the Fine-tuning and Det PL baselines and the oracle tracker.}
\label{tab:shift_exp}
\begin{tabular}{@{}llccccccc @{}}
\toprule \toprule
\multirow{2}{*}{\begin{tabular}[c]{@{}l@{}}\textbf{Setting}\\ Stage (+New Classes)\end{tabular}} & & \multicolumn{7}{c}{All Classes}                                                                                \\ \cmidrule(l){3-9} 
 & Method                                                                  & mMOTA          & mHOTA          & mIDF1          & MOTA& HOTA& IDF1&  mAP                       \\ \midrule \midrule

\textbf{M}$\rightarrow$\textbf{L}  Stage 0 (Pedestrian) &  & 53.7   &  46.1  &  54.4   &53.7&46.1&54.4&   43.0    \\
\midrule
\multirow{5}{*}{\begin{tabular}[c]{@{}l@{}}\textbf{M}$\rightarrow$\textbf{L}\\ Stage 1 (+Car) \end{tabular}}     & Fine-tuning    &  25.8   &  28.5    &   30.9      &24.2&40.5& 37.7&   25.3           \\ 
                                                                                                & Det PL      &  44.6        & 46.8         & 49.7        &44.2&47.2& 49.1&  45.6                              \\  
                                                                                                & COOLer     & \textbf{50.9}  & \textbf{50.9}  & \textbf{57.0} &\textbf{50.9}&\textbf{51.0}&\textbf{56.8} &  \textbf{45.7}            \\ 
                                                                                               &  \cellcolor{lightgray}Oracle       & \cellcolor{lightgray}53.7& \cellcolor{lightgray}51.8& \cellcolor{lightgray}58.7& \cellcolor{lightgray}53.7&  \cellcolor{lightgray}51.4&\cellcolor{lightgray}58.4&\cellcolor{lightgray}46.2            \\\midrule 
\multirow{5}{*}{\begin{tabular}[c]{@{}l@{}}\textbf{M}$\rightarrow$\textbf{L}\\ Stage 2 (+Truck)\end{tabular}}          & Fine-tuning    &   11.7      &  18.0        &   19.6      &2.7&16.9& 8.1  &   15.5                 \\ 
                                                                                                & Det PL      &     34.6      &   44.7       &    45.9 &33.5&43.0&  40.4   &    \textbf{44.8}            \\  
                                                                                                & COOLer      & \textbf{45.2}   &\textbf{ 51.5}   &\textbf{ 57.3}  &\textbf{48.2}&\textbf{50.6}&\textbf{56.3}&  \textbf{44.8}       \\ 
                                                                                                 & \cellcolor{lightgray}Oracle      & \cellcolor{lightgray}52.6 & \cellcolor{lightgray}53.5 &  \cellcolor{lightgray}60.7   &\cellcolor{lightgray}53.5 &\cellcolor{lightgray}51.8& \cellcolor{lightgray}58.9&  \cellcolor{lightgray}46.0      \\ \midrule
  \textbf{G}$\rightarrow$\textbf{S}  Stage 0 (General)  & &  50.8  & 53.1 & 60.0 &52.4&51.9&58.2& 46.0   \\ \midrule
\multirow{5}{*}{\begin{tabular}[c]{@{}l@{}}\textbf{G}$\rightarrow$\textbf{S}\\ Stage 1 (+Specific)\end{tabular}} & Fine-tuning          & 19.4          & 24.1          & 26.9        &4.8&18.1& 10.1 &    20.0               \\  
                                                                                               & Det PL   & 45.8          & 49.8          & 55.4          &48.7 &48.9&52.7&    \textbf{43.2}        \\  
                                                                                               & COOLer      &  \textbf{46.0} & \textbf{50.8} & \textbf{57.0} & \textbf{50.8}&\textbf{51.4}&\textbf{57.5}&   42.7                \\ 
                                                                                                &\cellcolor{lightgray}Oracle   &  \cellcolor{lightgray}48.8       &  \cellcolor{lightgray}51.1        & \cellcolor{lightgray}57.5         &  \cellcolor{lightgray}52.5&\cellcolor{lightgray}51.9&\cellcolor{lightgray}58.5&\cellcolor{lightgray}43.8       \\  
                                                                                               \midrule \midrule
                                                                                                \textbf{V}$\rightarrow$\textbf{B}$\rightarrow$\textbf{H}  Stage 0 (Vehicle) & &   47.2  & 52.1  & 57.4   &51.9&56.4&61.4& 45.2  \\ \midrule
\multirow{5}{*}{\begin{tabular}[c]{@{}l@{}}\textbf{V}$\rightarrow$\textbf{B}$\rightarrow$\textbf{H}\\ Stage 1 (+Bike)\end{tabular}}   & Fine-tuning          & 16.1          & 20.4          & 22.4         &5.9&20.5& 12.2&       16.6               \\  
                                                                                               & Det PL            & 39.4          & 47.1          & 51.0         & 41.4&47.4& 48.6&   42.0             \\  
                                                                                               & COOLer    &  \textbf{44.5} & \textbf{51.3} & \textbf{57.5} &\textbf{49.5}&\textbf{55.5}&\textbf{60.9}&  \textbf{42.3}  
                                                                                               \\ 
                                                                                                &\cellcolor{lightgray}Oracle   &  \cellcolor{lightgray}47.8        & \cellcolor{lightgray}52.1         & \cellcolor{lightgray}58.0        &\cellcolor{lightgray}51.5&\cellcolor{lightgray}55.5& \cellcolor{lightgray}60.8& \cellcolor{lightgray}44.2      \\  \midrule
\multirow{5}{*}{\begin{tabular}[c]{@{}l@{}}\textbf{V}$\rightarrow$\textbf{B}$\rightarrow$\textbf{H}\\ Stage 2  (+Human)\end{tabular}}  & Fine-tuning            & 8.9           & 7.7           & 9.1          &23.0&31.2&30.9 &          7.1          \\  
                                                                                               & Det PL            & 37.3          & 41.4          & 43.7        &39.8&40.9& 42.6 &     41.7              \\  
                                                                                               & COOLer    &\textbf{47.0} & \textbf{50.6} &  \textbf{57.5} &\textbf{50.7}&\textbf{51.3}&\textbf{57.7}& \textbf{42.2}     \\ 
                                                                                                &\cellcolor{lightgray}Oracle   &    \cellcolor{lightgray}48.8    &  \cellcolor{lightgray}51.1       &    \cellcolor{lightgray}57.5     &\cellcolor{lightgray}52.5&\cellcolor{lightgray}51.9&\cellcolor{lightgray}58.5&  \cellcolor{lightgray}43.8      \\  \bottomrule \bottomrule
\end{tabular}

\end{table*}

\subsection{Implementation Details}
COOLer's architecture is based on QDTrack with a ResNet-50~\cite{he2016deep} backbone. The model is optimized with SGD with momentum of 0.9 and weight decay of 1e-4. We train the network with 8 NVIDIA 2080Ti GPUs with a total batch size of 16. For all experiments, we train for 6 epochs in each incremental stage. The initial learning rate is 0.02 and decayed to 0.002 after 4 epochs, and to 0.0002 after 5. For \bdd experiments, the weight for the contrastive losses $\beta_1, \beta_2$ are 0.01; for \shift experiments, $\beta_1, \beta_2$ are 0.001. The hinge factor for the pushing loss $\Delta_{\mathrm{push}}$ is set to 15.0. For the prior distribution  $\mathcal{N}(\boldsymbol{\mu}_c, \mathrm{diag}(\boldsymbol{\sigma}_p^2))$ of the pulling loss, we use  $\boldsymbol{\sigma}_{p}=0.05\cdot\Vec{\mathbf{1}}$, where $\Vec{\mathbf{1}}$ is the unit vector. 
We select hyper-parameters from a grid search, and report sensitivity analysis in the supplement.
\begin{table*}[htbp]
\centering
\scriptsize
\setlength\tabcolsep{2pt}
\renewcommand\arraystretch{0.6}
\caption{\textbf{Ablation Study on Method Components.}  We ablate on the choice of pseudo-labels (PL) and contrastive (CT) loss for COOLer on \textbf{M}$\rightarrow$\textbf{L} setting on \bdd. We compare training with our pseudo-labels generated by the tracker (Track) and the pseudo-labels generated by the detector (Det). We also compare our contrastive loss (Ours) with the contrastive loss proposed in OWOD~\cite{joseph2021towards}.}
\label{tab:ab_ct_loss}
\begin{tabular}{@{}lccccccccc@{}}
\toprule
 
\multirow{2}{*}{\begin{tabular}[c]{@{}l@{}}\textbf{Setting}\\ Stage (+New Classes)\end{tabular}}                                        & \multicolumn{2}{c}{Components} & \multicolumn{7}{c}{All Classes}   \\\cmidrule(l){2-3} \cmidrule(l){4-10}                                                                               & PL & CT Loss                      & mMOTA          & mHOTA          & mIDF1      & MOTA & HOTA & IDF1    & mAP            \\ \midrule  
\multirow{4}{*}{\begin{tabular}[c]{@{}l@{}}\textbf{M}$\rightarrow$\textbf{L}\\ Stage 1 \\ (+Pedestrian)\end{tabular}} & Det      & \xmark                         & 46.7     &     49.5          &  59.2             & 56.3&53.9&61.8&      46.8          \\  
                                                                                            & Track      & \xmark                         & 54.1          & \textbf{52.6}          & \textbf{64.5} &62.5&59.3&70.3& 47.2  \\  
                                                                                            & Track      & OWOD~\cite{joseph2021towards} & 53.7 & 52.2 & 63.9          &62.1  &58.9&69.8&  47.2       \\   
                                                                                            & Track      & Ours                          & \textbf{54.2} & \textbf{52.6} & 64.3        & \textbf{62.7}& \textbf{59.5}&  \textbf{70.5}&  \textbf{47.4}         \\ \midrule
\multirow{4}{*}{\begin{tabular}[c]{@{}l@{}}\textbf{M}$\rightarrow$\textbf{L}\\ Stage 2 \\ (+Truck)\end{tabular}}      & Det      & \xmark                    & 34.9 & 47.1 &55.6 &57.3&55.5&65.1&   42.5          \\  
                                                                                            & Track     & \xmark                     & \textbf{43.4} & 49.1          & 59.5          &58.2 & 57.5 &68.3&\textbf{42.8}        \\  
                                                                                            & Track      & OWOD~\cite{joseph2021towards}  & 41.9          & 48.7 & 58.9 &57.6  &57.3&68.0& \textbf{42.8}  \\  
                                                                                            & Track      & Ours                      & 42.8          & \textbf{49.2} & \textbf{59.6} & \textbf{58.6} &\textbf{57.9}& \textbf{68.7}&  42.6   \\ \midrule
\multirow{4}{*}{\begin{tabular}[c]{@{}l@{}}\textbf{M}$\rightarrow$\textbf{L}\\ Stage 3 \\ (+Bus)\end{tabular}}        & Det      & \xmark                 & 14.1 &43.1 &49.1& 53.6&53.4&61.5& 40.9        \\\ 
                                                                                            & Track      & \xmark                      & 32.8          & 47.6          & 56.9          &  54.5 &56.3&66.8&  \textbf{41.9}      \\  
                                                                                            & Track      & OWOD~\cite{joseph2021towards}  &  33.1 &  47.5 & 56.9 &53.9&56.1&66.6& \textbf{41.9}  \\  
                                                                                            & Track      & Ours                       & \textbf{34.0} & \textbf{47.9} & \textbf{57.3} &\textbf{55.8} &\textbf{56.8}&\textbf{67.4}& \textbf{41.9}  \\ \midrule
\multirow{4}{*}{\begin{tabular}[c]{@{}l@{}}\textbf{M}$\rightarrow$\textbf{L}\\ Stage 4 \\ (+Bicycle)\end{tabular}}    & Det      & \xmark           & 3.6 & 37.6 & 42.4 &41.2&43.0&46.2&  34.3    \\  
                                                                                            & Track     & \xmark                  & 25.7          & 43.9          & 52.9          &50.8&55.1&65.4&  36.1  \\   
                                                                                            & Track   & OWOD~\cite{joseph2021towards} & 24.7 & 43.7 & 52.7 & 49.8&54.8&65.0& 36.1          \\  
                                                                                            & Track   & Ours                   & \textbf{28.6} & \textbf{44.4} & \textbf{53.9} &\textbf{53.2} &\textbf{55.7}&\textbf{66.1}&  \textbf{36.5}       \\ \bottomrule  
\end{tabular}

\end{table*}

\subsection{Experimental Results}
We compare our method to the above-mentioned baselines on the \bdd and SHIFT datasets. We evaluate the mAP for object detection, and representative tracking metrics for MOT. mMOTA, mHOTA, mIDF1 are averaged across category-specific metrics, while MOTA, HOTA, IDF1 are the overall metrics.

\noindent\textbf{\bdd.} Table~\ref{tab:bdd_exp} shows the results on the \bdd dataset. In the \textbf{M}$\rightarrow$\textbf{L} setting, we show results up to stage 4 (+Bicycle) due to space constraints, and report full results in the supplement. 
%
COOLer achieves the best tracking performance among all methods and in all settings. 
In (\textbf{M}$\rightarrow$\textbf{L}, Stage 4) COOLer has a noteworthy 48.0\%, and 25.0\%  mMOTA improvement compared to the Distillation and Det PL baselines, showing its effectiveness in class-incremental tracking.
Besides boosting the tracking performance, COOLer also improves continual object detection. Compared to the Distillation baseline based on the class-incremental object detector Faster-ILOD~\cite{peng2020faster}, COOLer achieves higher mAP thanks to our temporally-refined detection pseudo-labels.
COOLer obtains +1.4\% and +2.2\% mAP wrt. Distillation and Det PL in (\textbf{M}$\rightarrow$\textbf{L}, Stage 4), and +0.9\% and +1.4\% mAP wrt. Distillation and Det PL in (\textbf{V}$\rightarrow$\textbf{B}$\rightarrow$\textbf{H}, Stage 2).
%



\noindent\textbf{SHIFT.} 
We conduct experiments on the \shift dataset, and report the results in Tab.~\ref{tab:shift_exp}. The results confirm the findings and trends observed ob \bdd. COOLer consistently outperforms all other baselines across all tracking metrics on all classes, further showing the superiority and generality of our approach.


\subsection{Ablation Study}
We here ablate on method components and analyze qualitative results. 
In the supplement, we provide an additional analysis of the performance on old classes (model's rigidity) and new classes  (model's plasticity) under incremental stages.

\noindent\textbf{Ablation on Method Components.} 
We show the effectiveness of each proposed component of COOLer in Tab.~\ref{tab:ab_ct_loss}. 
We compare our detection pseudo-labels refined by the tracker (Track) vs. unrefined detection pseudo-labels from the object detector only (Det). Moreover, we analyze the effect of additional class-incremental contrastive losses, comparing ours (Ours) vs. OWOD's ~\cite{joseph2021towards} (OWOD).
Our components consistently improve over the baselines, and the improvement is more significant as more incremental stages are performed, suggesting that more stages pose a greater challenge in CIL for MOT. 
Notably, using the tracking pseudo-labels improves over all metrics, with 22.2\% mMOTA, 6.9\% mHOTA, 10.5\% mIDF1, and 1.8\% mAP at stage 4. Enabling the class-incremental contrastive loss further boosts 2.9\% mMOTA, 0.5\% mHOTA, and 1.0\% mIDF1, and 0.4\% mAP, highlighting the superiority of our contrastive loss.
The results confirm that COOLer can (i) utilize the tracker's temporal refinement to produce higher-quality labels for detection, (ii) better preserve the association performance thanks to the association pseudo-labels, and (iii) that our contrastive loss design outperforms OWOD's.
\begin{figure}[t]
\centering
\includegraphics[width=1.0\columnwidth]{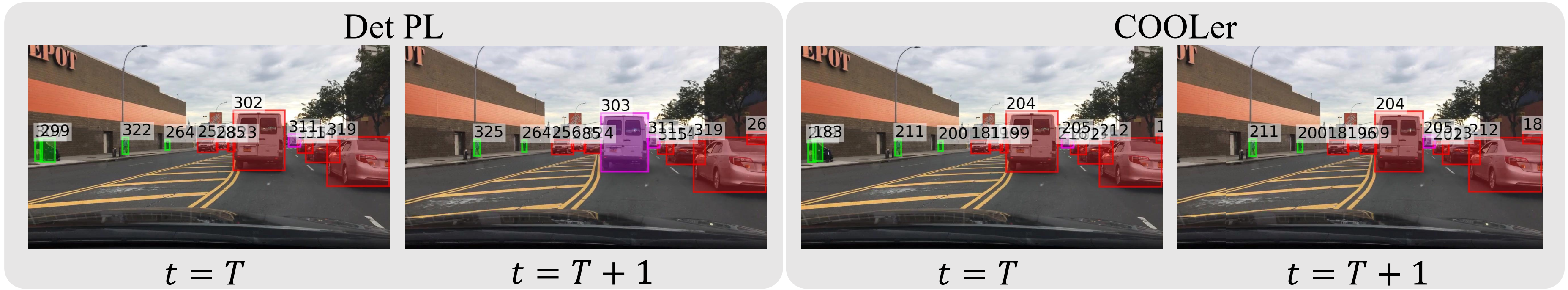} 
\caption{Qualitative Results of the Det PL baseline and COOLer on a validation video sequence of \bdd in the fourth step of the \textbf{M}$\rightarrow$\textbf{L} setting (+Bicycle). Different bounding box colors represent different classes, and the number above the bound box denotes the instance ID. Best viewed in color with zoom.} 
\label{fig:det_track_pl}
\vspace{-2mm}
\end{figure}
\noindent\textbf{Qualitative Comparison.}
Fig.~\ref{fig:det_track_pl} shows that the Det PL baseline would suffer from ID switches of the car (red) in the middle, due to the misclassification of it as a bus (purple) in the second frame. It can also not associate the pedestrians beside the pole across two frames (ID 322 switches to ID 325). Nevertheless, COOLer can both correctly classify the car and associate the pedestrians. This demonstrates that COOLer can better retain the knowledge of associating objects of the old classes while reducing misclassifications.
\section{Conclusion}
Our work is the first to address continual learning for MOT, a practical problem as MOT datasets are expensive to collect.
We introduce COOLer, the first comprehensive approach to class-incremental learning for multiple object tracking. 
COOLer adopts a continual pseudo-label generation strategy for tracking that leverages the previous tracker to generate association pseudo-labels and temporally-refine detection pseudo-labels, while introducing class-incremental instance representation learning to further improve the tracking performance.
%
Experimental results demonstrate that COOLer overcomes the drawbacks of detection-oriented methods, improving both detection and association performance.
Although highly effective in the proposed setting, COOLer assumes that instances from the previous classes are present in the new training data. 
We believe experience replay to be a possible solution to this limitation, and we leave its exploration to future work.
%
%
We hope our work can stimulate future research in this challenging yet practical direction.
%

\section{Acknowledgments}
This work was funded by the Max Planck ETH Center for Learning Systems.
%
%
%
%
\bibliographystyle{splncs04}
\bibliography{egbib}

\begin{thebibliography}{10}
\providecommand{\url}[1]{\texttt{#1}}
\providecommand{\urlprefix}{URL }
\providecommand{\doi}[1]{https://doi.org/#1}

\bibitem{aharon2022bot}
Aharon, N., Orfaig, R., Bobrovsky, B.Z.: Bot-sort: Robust associations
  multi-pedestrian tracking. arXiv preprint arXiv:2206.14651  (2022)

\bibitem{bernardin2008evaluating}
Bernardin, K., Stiefelhagen, R.: Evaluating multiple object tracking
  performance: the clear mot metrics. EURASIP Journal on Image and Video
  Processing  \textbf{2008},  1--10 (2008)

\bibitem{cha2021co2l}
Cha, H., Lee, J., Shin, J.: Co2l: Contrastive continual learning. In:
  Proceedings of the IEEE/CVF International Conference on Computer Vision. pp.
  9516--9525 (2021)

\bibitem{chen2020simple}
Chen, T., Kornblith, S., Norouzi, M., Hinton, G.: A simple framework for
  contrastive learning of visual representations. In: International conference
  on machine learning. pp. 1597--1607. PMLR (2020)

\bibitem{chen2018lifelong}
Chen, Z., Liu, B.: Lifelong machine learning. Synthesis Lectures on Artificial
  Intelligence and Machine Learning  \textbf{12}(3),  1--207 (2018)

\bibitem{dave2020tao}
Dave, A., Khurana, T., Tokmakov, P., Schmid, C., Ramanan, D.: Tao: A
  large-scale benchmark for tracking any object. In: European conference on
  computer vision. pp. 436--454. Springer (2020)

\bibitem{dendorfer2020mot20}
Dendorfer, P., Rezatofighi, H., Milan, A., Shi, J., Cremers, D., Reid, I.,
  Roth, S., Schindler, K., Leal-Taix{\'e}, L.: Mot20: A benchmark for multi
  object tracking in crowded scenes. arXiv preprint arXiv:2003.09003  (2020)

\bibitem{du2023strongsort}
Du, Y., Zhao, Z., Song, Y., Zhao, Y., Su, F., Gong, T., Meng, H.: Strongsort:
  Make deepsort great again. IEEE Transactions on Multimedia  (2023)

\bibitem{fan2018unsupervised}
Fan, H., Zheng, L., Yan, C., Yang, Y.: Unsupervised person re-identification:
  Clustering and fine-tuning. ACM Transactions on Multimedia Computing,
  Communications, and Applications (TOMM)  \textbf{14}(4),  1--18 (2018)

\bibitem{fan2022normalization}
Fan, Q., Segu, M., Tai, Y.W., Yu, F., Tang, C.K., Schiele, B., Dai, D.:
  Normalization perturbation: A simple domain generalization method for
  real-world domain shifts. arXiv preprint arXiv:2211.04393  (2022)

\bibitem{fan2022towards}
Fan, Q., Segu, M., Tai, Y.W., Yu, F., Tang, C.K., Schiele, B., Dai, D.: Towards
  robust object detection invariant to real-world domain shifts. In: The
  Eleventh International Conference on Learning Representations (2022)

\bibitem{he2020momentum}
He, K., Fan, H., Wu, Y., Xie, S., Girshick, R.: Momentum contrast for
  unsupervised visual representation learning. In: Proceedings of the IEEE/CVF
  Conference on Computer Vision and Pattern Recognition. pp. 9729--9738 (2020)

\bibitem{he2016deep}
He, K., Zhang, X., Ren, S., Sun, J.: Deep residual learning for image
  recognition. In: Proceedings of the IEEE conference on computer vision and
  pattern recognition. pp. 770--778 (2016)

\bibitem{joseph2021towards}
Joseph, K., Khan, S., Khan, F.S., Balasubramanian, V.N.: Towards open world
  object detection. In: Proceedings of the IEEE/CVF Conference on Computer
  Vision and Pattern Recognition. pp. 5830--5840 (2021)

\bibitem{karthik2020simple}
Karthik, S., Prabhu, A., Gandhi, V.: Simple unsupervised multi-object tracking.
  arXiv preprint arXiv:2006.02609  (2020)

\bibitem{kirkpatrick2017overcoming}
Kirkpatrick, J., Pascanu, R., Rabinowitz, N., Veness, J., Desjardins, G., Rusu,
  A.A., Milan, K., Quan, J., Ramalho, T., Grabska-Barwinska, A., et~al.:
  Overcoming catastrophic forgetting in neural networks. Proceedings of the
  national academy of sciences  \textbf{114}(13),  3521--3526 (2017)

\bibitem{li2019unsupervised}
Li, M., Zhu, X., Gong, S.: Unsupervised tracklet person re-identification. IEEE
  transactions on pattern analysis and machine intelligence  \textbf{42}(7),
  1770--1782 (2019)

\bibitem{li2017learning}
Li, Z., Hoiem, D.: Learning without forgetting. IEEE transactions on pattern
  analysis and machine intelligence  \textbf{40}(12),  2935--2947 (2017)

\bibitem{liu2023continual}
Liu, Y., Schiele, B., Vedaldi, A., Rupprecht, C.: Continual detection
  transformer for incremental object detection. In: Proceedings of the IEEE/CVF
  Conference on Computer Vision and Pattern Recognition. pp. 23799--23808
  (2023)

\bibitem{mai2021supervised}
Mai, Z., Li, R., Kim, H., Sanner, S.: Supervised contrastive replay: Revisiting
  the nearest class mean classifier in online class-incremental continual
  learning. In: Proceedings of the IEEE/CVF Conference on Computer Vision and
  Pattern Recognition. pp. 3589--3599 (2021)

\bibitem{mallya2018packnet}
Mallya, A., Lazebnik, S.: Packnet: Adding multiple tasks to a single network by
  iterative pruning. In: Proceedings of the IEEE conference on Computer Vision
  and Pattern Recognition. pp. 7765--7773 (2018)

\bibitem{mccloskey1989catastrophic}
McCloskey, M., Cohen, N.J.: Catastrophic interference in connectionist
  networks: The sequential learning problem. In: Psychology of learning and
  motivation, vol.~24, pp. 109--165. Elsevier (1989)

\bibitem{pang2021quasi}
Pang, J., Qiu, L., Li, X., Chen, H., Li, Q., Darrell, T., Yu, F.: Quasi-dense
  similarity learning for multiple object tracking. In: Proceedings of the
  IEEE/CVF conference on computer vision and pattern recognition. pp. 164--173
  (2021)

\bibitem{peng2020faster}
Peng, C., Zhao, K., Lovell, B.C.: Faster ilod: Incremental learning for object
  detectors based on faster rcnn. Pattern Recognition Letters  \textbf{140},
  109--115 (2020)

\bibitem{ramanan2003finding}
Ramanan, D., Forsyth, D.A.: Finding and tracking people from the bottom up. In:
  2003 IEEE Computer Society Conference on Computer Vision and Pattern
  Recognition, 2003. Proceedings. vol.~2, pp. II--II. IEEE (2003)

\bibitem{rebuffi2017icarl}
Rebuffi, S.A., Kolesnikov, A., Sperl, G., Lampert, C.H.: icarl: Incremental
  classifier and representation learning. In: Proceedings of the IEEE
  conference on Computer Vision and Pattern Recognition. pp. 2001--2010 (2017)

\bibitem{ren2015faster}
Ren, S., He, K., Girshick, R., Sun, J.: Faster r-cnn: Towards real-time object
  detection with region proposal networks. Advances in neural information
  processing systems  \textbf{28} (2015)

\bibitem{rusu2016progressive}
Rusu, A.A., Rabinowitz, N.C., Desjardins, G., Soyer, H., Kirkpatrick, J.,
  Kavukcuoglu, K., Pascanu, R., Hadsell, R.: Progressive neural networks. arXiv
  preprint arXiv:1606.04671  (2016)

\bibitem{segu2023darth}
Segu, M., Schiele, B., Yu, F.: Darth: Holistic test-time adaptation for
  multiple object tracking. In: Proceedings of the IEEE/CVF International
  Conference on Computer Vision (2023)

\bibitem{shmelkov2017incremental}
Shmelkov, K., Schmid, C., Alahari, K.: Incremental learning of object detectors
  without catastrophic forgetting. In: Proceedings of the IEEE international
  conference on computer vision. pp. 3400--3409 (2017)

\bibitem{shift2022}
Sun, T., Segu, M., Postels, J., Wang, Y., Van~Gool, L., Schiele, B., Tombari,
  F., Yu, F.: {SHIFT:} a synthetic driving dataset for continuous multi-task
  domain adaptation. In: Proceedings of the IEEE/CVF Conference on Computer
  Vision and Pattern Recognition (CVPR). pp. 21371--21382 (June 2022)

\bibitem{wang2022smiletrack}
Wang, Y.H.: Smiletrack: Similarity learning for multiple object tracking. arXiv
  preprint arXiv:2211.08824  (2022)

\bibitem{Wojke2017simple}
Wojke, N., Bewley, A., Paulus, D.: Simple online and realtime tracking with a
  deep association metric. In: 2017 IEEE International Conference on Image
  Processing (ICIP). pp. 3645--3649. IEEE (2017).
  \doi{10.1109/ICIP.2017.8296962}

\bibitem{wu2020tracklet}
Wu, G., Zhu, X., Gong, S.: Tracklet self-supervised learning for unsupervised
  person re-identification. In: Proceedings of the AAAI Conference on
  Artificial Intelligence. vol.~34, pp. 12362--12369 (2020)

\bibitem{xie2022general}
Xie, J., Yan, S., He, X.: General incremental learning with domain-aware
  categorical representations. In: Proceedings of the IEEE/CVF Conference on
  Computer Vision and Pattern Recognition. pp. 14351--14360 (2022)

\bibitem{yu2020bdd100k}
Yu, F., Chen, H., Wang, X., Xian, W., Chen, Y., Liu, F., Madhavan, V., Darrell,
  T.: Bdd100k: A diverse driving dataset for heterogeneous multitask learning.
  In: Proceedings of the IEEE/CVF conference on computer vision and pattern
  recognition. pp. 2636--2645 (2020)

\bibitem{zhang2022bytetrack}
Zhang, Y., Sun, P., Jiang, Y., Yu, D., Weng, F., Yuan, Z., Luo, P., Liu, W.,
  Wang, X.: Bytetrack: Multi-object tracking by associating every detection
  box. In: European Conference on Computer Vision. pp. 1--21. Springer (2022)

\bibitem{zheng2021contrast}
Zheng, K., Chen, C.: Contrast r-cnn for continual learning in object detection.
  arXiv preprint arXiv:2108.04224  (2021)

\bibitem{zhou2020lifelong}
Zhou, W., Chang, S., Sosa, N., Hamann, H., Cox, D.: Lifelong object detection.
  arXiv preprint arXiv:2009.01129  (2020)

\bibitem{zhu2020deformable}
Zhu, X., Su, W., Lu, L., Li, B., Wang, X., Dai, J.: Deformable detr: Deformable
  transformers for end-to-end object detection. arXiv preprint arXiv:2010.04159
   (2020)

\end{thebibliography}
\newpage
\centerline{\large{\textbf{Supplementary Material}}}
\vskip 0.1in
In this supplementary material, we provide additional implementation details as well as experiment results and analysis regarding our proposed COntrastive- and cOntinual-Learning-based tracker (COOLer) for class-incremental learning for appearance-based multiple object tracking (MOT), that we could not discuss in the main paper due to space constraints.
\appendix
\section{Details of the Memory Queue} \label{sup:sec:memory}
To estimate the class mean and standard deviation prototype vectors $\boldsymbol{\mu}_c, \boldsymbol{\sigma}_c$, we keep a memory queue of size $N_{\mathrm{queue}}$ for each class to store exemplar embedding vectors. The memory queue is updated during training to handle the changing parameters of the network and the intra-class variability in the training data. 
At each training step, we first update the memory queue by sampling at most $N_{\mathrm{batch}}$ embedding vectors for each class in the current batch and push them to the memory queue.
Then, we compute the mean $\hat{\boldsymbol{\mu}}_c$ and the standard deviation $\hat{\boldsymbol{\sigma}}_c$ of the embedding vectors in the memory queue for each class, given there are already more than $N_{\mathrm{class}}$ exemplar vectors for that class. 
Finally, we update the mean and standard deviation prototype vectors through Polyak averaging:
\begin{equation}
\boldsymbol{\mu}_c = \eta \boldsymbol{\mu}_c + (1-\eta) \hat{\boldsymbol{\mu}}_c,
\boldsymbol{\sigma}_c = \eta \boldsymbol{\sigma}_c + (1-\eta) \hat{\boldsymbol{\sigma}}_c,
\end{equation}
where $\eta$ is the Polyak factor.
The Polyak averaging ensures a more smooth update of the prototype vectors. 
%
Note that the memory queue, $\hat{\boldsymbol{\mu}}_c$, and  $\hat{\boldsymbol{\sigma}}_c$ are copied together with the network weights in later incremental steps, which helps preserve the embedding distribution of the old classes. In our implementation, we use $N_{\mathrm{queue}} = 1000$, $N_{\mathrm{batch}}=2$, $N_{\mathrm{class}}=100$, and $\eta=0.999$.

\section{Additional Implementation Details}
In each incremental stage, we first initialize the new tracking model from the same weights as the old one. Its last fully connected layers of the classification head and the bounding box regression head of the object detector are extended to incorporate the new classes.  As the object detector and the similarity network are jointly optimized, we apply the contrastive loss on the Re-ID embedding of the similarity network. For objects of the new classes, their embeddings may be unstable at the beginning of the training, and hence we enable the contrastive learning for the new classes only after the first epoch of training.
For the training data, we pick all video sequences in the training split where at least one object from the new classes is presented, and all other video sequences are discarded. 
As tracking datasets are organized as video sequences, we keep all frames of the video sequences, even for frames with no object from the new classes presented.  
We then keep the labels of the new classes and remove all annotations for the old classes.
We use the whole validation split to evaluate trackers and only evaluate the classes that the model has learned.

\section{Performance of Old and New Classes}
In each incremental step, we additionally compute the average performance of all old classes to evaluate the ability to retain old knowledge (the model rigidity) and the average performance of all new classes to evaluate the ability to acquire new knowledge (the model plasticity). The results on the \bdd dataset is shown in Tab.~\ref{tab:bdd_exp_full}.
We can see that COOLer also achieves the best rigidity-plasticity trade-off when compared to the oracle model. In Stage 1 of the \textbf{G}$\rightarrow$\textbf{S} setting, COOLer only loses 1.5\% and 2.1\% mMOTA for the old and the new classes compared to the oracle. Even though the Distillation baseline has better performance on the old classes, it fails to learn the new classes by losing 189.1\% mMOTA. This suggests that distillation-based methods focus too much on the model's rigidity rather than plasticity in class-incremental learning for MOT. On the other hand, the Fine-tuning and Det PL baselines focus too much on the model's plasticity rather than rigidity. The results on the \shift dataset is shown in Tab.~\ref{tab:shift_exp_full}, we observe similar trends that the other baseline suffers from retaining old knowledge, while COOLer can keep old knowledge effectively while learning new knowledge.  In Tab.~\ref{tab:ab_full},  we also compute the performance of old and new classes in the ablation study on method components. We can see that with our proposed pseudo-labels generated by the tracker and the class-incremental contrastive loss, the tracker achieves the best performance on both old and new classes especially in later incremental steps, further demonstrating the superiority of our  method.

\begin{table*}[htbp]
\scriptsize 
\setlength\tabcolsep{1.6pt}
\caption{Performance of old and new classes on \bdd. We conduct experiments on \textbf{M}$\rightarrow$\textbf{L}, \textbf{G}$\rightarrow$\textbf{S} and \textbf{V}$\rightarrow$\textbf{B}$\rightarrow$\textbf{H} settings.
We compare COOLer with the Fine-tuning, Distillation, Det PL baselines and the oracle tracker.  }
\label{tab:bdd_exp_full}
\begin{subtable}[htbp]{1.0\textwidth} 
\centering
\caption{Mean Metrics }
\label{tab:bdd_exp_full_a}
\begin{tabular}{@{}llcccc|cccc@{}}
\toprule \toprule
\multirow{2}{*}{\begin{tabular}[c]{@{}l@{}}\textbf{Setting}\\ Stage (+New Classes)\end{tabular}}                                              &                                                    & \multicolumn{4}{c}{Old Classes}                                               & \multicolumn{4}{c}{New Classes}                                               \\ \cmidrule(l){3-10} 
&Method                                                                                &\tiny{mMOTA}          &  \tiny{mHOTA}          & \tiny{mIDF1}          & \tiny{mAP}                & \tiny{mMOTA}          & \tiny{mHOTA}          & \tiny{mIDF1}          & \tiny{mAP}          \\ \midrule \midrule

\textbf{M}$\rightarrow$\textbf{L}  Stage 0 (Car) &    & - & - &- & - &  67.6 & 62.1 & 73.3 & 58.7\\
\midrule
\multirow{5}{*}{\begin{tabular}[c]{@{}l@{}}\textbf{M}$\rightarrow$\textbf{L}\\ Stage 1 (+Pedestrian) \end{tabular}}     & Fine-tuning            & 0.0           & 0.0           & 0.0           & 0.0               &   31.3        &   43.6       &  55.4   &  \textbf{39.8}         \\ 
                                                                                                & Distillation              & \textbf{68.0}          & \textbf{62.1}         & \textbf{73.4}        & \textbf{58.3}                    & 24.9         & 41.3          &   52.6        &   36.0                 \\  
                                                                                                & Det PL                    &  60.1        & 55.7          &    63.0       &  55.1          & 33.3         &  \textbf{43.4}         & 55.4          &  38.4                \\  
                                                                                                & COOLer        & 66.0 & 61.8 & 72.8 & 57.1   & \textbf{42.5} & \textbf{43.4} &  \textbf{55.8}        & 37.7           \\ 
                                                                                               &  \cellcolor{lightgray}Oracle       &  \cellcolor{lightgray}68.2 & \cellcolor{lightgray}62.4 & \cellcolor{lightgray}73.8 & \cellcolor{lightgray}58.6   & \cellcolor{lightgray}46.6 & \cellcolor{lightgray}44.7 & \cellcolor{lightgray}57.9         & \cellcolor{lightgray}38.0           \\\midrule 
\multirow{5}{*}{\begin{tabular}[c]{@{}l@{}}\textbf{M}$\rightarrow$\textbf{L}\\ Stage 2 (+Truck)\end{tabular}}          & Fine-tuning          & 0.0           & 0.0           & 0.0           & 0.0            &  -34.5        & 38.4         &  42.0         &  35.0                \\ 
                                                                                                & Distillation          & \textbf{52.3}          &  \textbf{51.8}         &  \textbf{63.7}         &  \textbf{46.9}                & -22.4         & 39.7       &   44.6        &  34.4                   \\  
                                                                                                & Det PL           &  49.1        & 49.3          &  58.8        &  45.3         &  6.4             &  42.9         & 49.3          & \textbf{36.9}            \\  
                                                                                                & COOLer       & 50.8 &  \textbf{51.8}& 63.3 & 46.1 & \textbf{26.7} & \textbf{43.9} & \textbf{52.1}  & 35.7          \\ 
                                                                                                 & \cellcolor{lightgray}Oracle      &  \cellcolor{lightgray}57.4 & \cellcolor{lightgray}53.6 & \cellcolor{lightgray}65.9 & \cellcolor{lightgray}48.3 & \cellcolor{lightgray}34.5& \cellcolor{lightgray}45.4 & \cellcolor{lightgray}54.5 & \cellcolor{lightgray}38.3          \\ \midrule
\multirow{5}{*}{\begin{tabular}[c]{@{}l@{}}\textbf{M}$\rightarrow$\textbf{L}\\ Stage 3  (+Bus)\end{tabular}}            & Fine-tuning                  & 0.0           & 0.0           & 0.0           & 0.0           &   -96.0        &  36.8                  & 37.3          &   39.7        \\  
                                                                                                & Distillation                 &   \textbf{41.5}         &    47.1        &    57.4      &  \textbf{42.0}         &  -169.5       &   32.3          & 31.0          &   37.5         \\  
                                                                                                & Det PL            & 36.5           & 44.1           &     51.2      &   40.8         &   -53.2        &  40.0          & 42.8          & 41.3  \\  
                                                                                                & COOLer        &  41.2 & \textbf{48.5} &  \textbf{58.6} & 41.7  & \textbf{12.4}  &  \textbf{46.2} &  \textbf{53.3}  &  \textbf{42.4}    \\
                                                                                                & \cellcolor{lightgray}Oracle      &  \cellcolor{lightgray}49.8  & \cellcolor{lightgray}50.8 & \cellcolor{lightgray}62.1  & \cellcolor{lightgray}45.0  & \cellcolor{lightgray}32.1  & \cellcolor{lightgray}47.7  &\cellcolor{lightgray}57.3   & \cellcolor{lightgray}42.9     \\\midrule
\multirow{5}{*}{\begin{tabular}[c]{@{}l@{}}\textbf{M}$\rightarrow$\textbf{L}\\ Stage 4  (+Bicycle)\end{tabular}}        & Fine-tuning              & 0.0           & 0.0           & 0.0           & 0.0           &     -80.0   &   28.1        &  32.3          &  \textbf{23.0}        \\ 
                                                                                                & Distillation                               &  18.2          &   44.2       &     52.8       &   39.8       &    -169.6      &  23.7         & 24.5 &16.3          \\ 
                                                                                                & Det PL              &  18.8         &  39.6          &  44.4          &  37.9         &  -57.6       &  29.7          &    34.5       &  20.3 \\  
                                                                                                & COOLer        & \textbf{32.2} & \textbf{47.1}  & \textbf{56.7}  & \textbf{40.7}  & \textbf{14.1}  & \textbf{33.5}  & \textbf{42.6}  & 19.6          \\
                                                                                                   & \cellcolor{lightgray}Oracle       &  \cellcolor{lightgray}45.4 &  \cellcolor{lightgray}50.1 & \cellcolor{lightgray}60.9 & \cellcolor{lightgray}44.5  & \cellcolor{lightgray}25.0  &\cellcolor{lightgray}35.5  & \cellcolor{lightgray}46.3  & \cellcolor{lightgray}21.8          \\\midrule \midrule
 \textbf{G}$\rightarrow$\textbf{S}  Stage 0 (General)  &    & - & - &- & - &  45.6  &  50.3 &  61.1&  44.6\\
\midrule                                                                                                 
\multirow{5}{*}{\begin{tabular}[c]{@{}l@{}}\textbf{G}$\rightarrow$\textbf{S}\\ Stage 1 (+Specific)\end{tabular}} & Fine-tuning                        & 0.0           & 0.0           & 0.0           & 0.0          &  -49.4        &   23.5        &  28.2         &     16.8     \\  
                                                                                                & Distillation          &  \textbf{47.8}         &  49.5         &   \textbf{60.5}        &   \textbf{43.9}        &   -113.7     &    21.4     &  24.4         & 15.2         \\  
                                                                                                & Det PL                    &  39.8      &  44.7         &   52.4        &  38.9         &   -27.8       &    25.1       &     30.6      & \textbf{17.0} \\  
                                                                                                & COOLer             & 43.9 &  
                                                                                            \textbf{49.8}& \textbf{60.5} & 42.8 & \textbf{13.4} & \textbf{27.3} & \textbf{35.9} &  \textbf{17.0}         \\ 
                                                                                                & \cellcolor{lightgray}Oracle             &\cellcolor{lightgray}45.4  &\cellcolor{lightgray}50.1 &\cellcolor{lightgray}60.9 & \cellcolor{lightgray}44.5 &\cellcolor{lightgray}15.5  & \cellcolor{lightgray}27.7  & \cellcolor{lightgray}37.2   &  \cellcolor{lightgray}17.4     \\ \midrule \midrule
                                                                                               \textbf{V}$\rightarrow$\textbf{B}$\rightarrow$\textbf{H}  Stage 0 (Vehicle) &   & - & - &- & - &  33.1    & 39.2  & 46.8 &  35.0 \\
\midrule
\multirow{5}{*}{\begin{tabular}[c]{@{}l@{}}\textbf{V}$\rightarrow$\textbf{B}$\rightarrow$\textbf{H}\\ Stage 1 (+Bike)\end{tabular}}   & Fine-tuning                  & 0.0           & 0.0           & 0.0           & 0.0       &  -81.6        &     29.1      &  33.7        &   \textbf{21.9}       \\ 
                                                                                                & Distillation           &  \textbf{36.5}         &  \textbf{39.0}         &    \textbf{46.4}       &    \textbf{34.9}       &  -164.5       &  24.3        &  26.1         &  17.5         \\  
                                                                                                & Det PL                 &   26.2                   &  31.9         &   36.4        &     29.6    &     -62.3      &   30.1   &   35.1  &  20.9\\  
                                                                                                & COOLer         & 32.4 & 38.2 & 45.3 & 33.5 & \textbf{8.6}  & \textbf{34.2} & \textbf{44.0} &  20.3         \\ 
                                                                                                & \cellcolor{lightgray}Oracle            & \cellcolor{lightgray}29.2  & \cellcolor{lightgray}38.9 &  \cellcolor{lightgray}46.4& \cellcolor{lightgray}35.0 &  \cellcolor{lightgray}24.6 &  \cellcolor{lightgray}37.8& \cellcolor{lightgray}50.6& \cellcolor{lightgray}22.4          \\ \midrule
\multirow{5}{*}{\begin{tabular}[c]{@{}l@{}}\textbf{V}$\rightarrow$\textbf{B}$\rightarrow$\textbf{H}\\ Stage 2 (+Human)\end{tabular}}  & Fine-tuning                & 0.0           & 0.0           & 0.0           &  0.0  &   29.5        &  39.9 &     52.3      & \textbf{33.0} \\  
                                                                                                & Distillation            &   15.3        &        36.1   &  43.3         &   \textbf{28.4}        &  13.5         &  35.2         &  45.5        &  26.5         \\  
                                                                                                & Det PL        &     10.0      & 32.8          &    37.9       &    25.9       &    32.0      &    \textbf{39.8}      &    52.4      & 31.8\\  
                                                                                                & COOLer      & \textbf{23.8} &  \textbf{36.9} & \textbf{44.9}  &   28.3      & \textbf{37.2} &  39.4    & \textbf{52.5} &   30.4        \\ 
                                                                                                 &\cellcolor{lightgray}Oracle   &\cellcolor{lightgray}27.6 &\cellcolor{lightgray}38.5  & \cellcolor{lightgray}47.8 &\cellcolor{lightgray}30.8  &\cellcolor{lightgray}38.8  & \cellcolor{lightgray}40.0     &  \cellcolor{lightgray}53.2& \cellcolor{lightgray}31.4          \\ \bottomrule \bottomrule 
\end{tabular}
\end{subtable}
\end{table*}
\begin{table*}[htbp]
\scriptsize 
\centering
\setlength\tabcolsep{5.3pt}
\ContinuedFloat
\begin{subtable}[htbp]{1.0\textwidth}
\caption{Overall Metrics}
\centering
\label{tab:bdd_exp_full_b}
\begin{tabular}{@{}llccc|ccc@{}}
\toprule \toprule
\multirow{2}{*}{\begin{tabular}[c]{@{}l@{}}\textbf{Setting}\\ Stage (+New Classes)\end{tabular}}                                              &                                                    & \multicolumn{3}{c}{Old Classes}                                               & \multicolumn{3}{c}{New Classes}                                               \\ \cmidrule(l){3-8} 
&Method                                                                                & MOTA          &   HOTA          & IDF1               & MOTA          &   HOTA          & IDF1                 \\ \midrule \midrule

\textbf{M}$\rightarrow$\textbf{L}  Stage 0 (Car) &    & - & - &- & 67.6  &62.1&73.3\\
\midrule
\multirow{5}{*}{\begin{tabular}[c]{@{}l@{}}\textbf{M}$\rightarrow$\textbf{L}\\ Stage 1 (+Pedestrian) \end{tabular}}     & Fine-tuning         &0.0&0.0&0.0&31.3&43.6&55.4    \\ 
                                                                                                & Distillation          &\textbf{68.0}&\textbf{62.1}&\textbf{73.4}&24.9&41.3&52.6            \\  
                                                                                                & Det PL                &60.1&55.7&63.0&33.3&\textbf{43.4}&55.4       \\  
                                                                                                & COOLer      &66.0&61.8&72.8&\textbf{42.5}&\textbf{43.4}&\textbf{55.8}           \\ 
                                                                                               &  \cellcolor{lightgray}Oracle       &\cellcolor{lightgray}68.2&\cellcolor{lightgray}62.4&\cellcolor{lightgray}73.8&\cellcolor{lightgray}46.6&\cellcolor{lightgray}44.7&\cellcolor{lightgray}57.9       \\\midrule 
\multirow{5}{*}{\begin{tabular}[c]{@{}l@{}}\textbf{M}$\rightarrow$\textbf{L}\\ Stage 2 (+Truck)\end{tabular}}          & Fine-tuning    &0.0&0.0&0.0&-34.5&38.4&42.0            \\ 
                                                                                                & Distillation        &\textbf{62.3}&58.2&69.5&-22.4&39.7&44.6               \\  
                                                                                                & Det PL         &60.8&56.5&66.3&6.4&42.9&49.3      \\  
                                                                                                & COOLer       &60.8&\textbf{58.7}&\textbf{69.6}&\textbf{26.7}&\textbf{43.9}&\textbf{52.1}         \\ 
                                                                                                 & \cellcolor{lightgray}Oracle  &\cellcolor{lightgray}65.1&\cellcolor{lightgray}59.9&\cellcolor{lightgray}71.5&\cellcolor{lightgray}34.5&\cellcolor{lightgray}45.4&\cellcolor{lightgray}54.5        \\ \midrule
\multirow{5}{*}{\begin{tabular}[c]{@{}l@{}}\textbf{M}$\rightarrow$\textbf{L}\\ Stage 3  (+Bus)\end{tabular}}            & Fine-tuning               &0.0&0.0&0.0&-96.0&36.8&37.3        \\  
                                                                                                & Distillation            &\textbf{58.8}&56.3&67.3&-169.5&32.3&31.0         \\  
                                                                                                & Det PL      &55.9&53.9&62.1&-53.2&40.0&42.8\\  
                                                                                                & COOLer      &56.8&\textbf{57.1}&\textbf{67.8}&\textbf{12.4}&\textbf{46.2}&\textbf{53.3}  \\
                                                                                                & \cellcolor{lightgray}Oracle &\cellcolor{lightgray}63.2&\cellcolor{lightgray}58.9&\cellcolor{lightgray}70.4&\cellcolor{lightgray}32.1&\cellcolor{lightgray}47.7&\cellcolor{lightgray}57.3     \\\midrule
\multirow{5}{*}{\begin{tabular}[c]{@{}l@{}}\textbf{M}$\rightarrow$\textbf{L}\\ Stage 4  (+Bicycle)\end{tabular}}        & Fine-tuning           &0.0&0.0&0.0&-80.0&28.1&32.3    \\ 
                                                                                                & Distillation                         &\textbf{54.0}&53.8&64.4&-169.6&23.7&24.5      \\ 
                                                                                                & Det PL         &42.2&43.2&46.4&-57.6&29.7&34.5 \\  
                                                                                                & COOLer  &53.5&\textbf{55.9}&\textbf{66.3}&\textbf{14.1}&\textbf{33.5}&\textbf{42.6}   \\
                                                                                                   & \cellcolor{lightgray}Oracle  &\cellcolor{lightgray}62.5&\cellcolor{lightgray}58.7&\cellcolor{lightgray}70.2&\cellcolor{lightgray}25.0&\cellcolor{lightgray}35.5&\cellcolor{lightgray}46.3         \\\midrule \midrule
 \textbf{G}$\rightarrow$\textbf{S}  Stage 0 (General)  &     & - &- & - & 62.4 &59.0& 70.2\\
\midrule                                                                                                 
\multirow{5}{*}{\begin{tabular}[c]{@{}l@{}}\textbf{G}$\rightarrow$\textbf{S}\\ Stage 1 (+Specific)\end{tabular}} & Fine-tuning                     &0.0&0.0&0.0&-30.0&31.5&38.3     \\  
                                                                                                & Distillation       &\textbf{62.1}&58.1&69.3&-78.8&28.6&33.3    \\  
                                                                                                & Det PL       &55.3&52.4&60.0&-11.1&33.2&41.4 \\  
                                                                                                & COOLer        &61.2&\textbf{58.5}&\textbf{69.7}&\textbf{21.0}&\textbf{35.2}&\textbf{45.7}        \\ 
                                                                                                & \cellcolor{lightgray}Oracle     &\cellcolor{lightgray}62.5&\cellcolor{lightgray}58.7&\cellcolor{lightgray}70.2&\cellcolor{lightgray}25.1&\cellcolor{lightgray}34.5&\cellcolor{lightgray}46.1       \\ \midrule \midrule
                                                                                               \textbf{V}$\rightarrow$\textbf{B}$\rightarrow$\textbf{H}  Stage 0 (Vehicle) &   & - &- & -  &65.1&61.0&72.3 \\
\midrule
\multirow{5}{*}{\begin{tabular}[c]{@{}l@{}}\textbf{V}$\rightarrow$\textbf{B}$\rightarrow$\textbf{H}\\ Stage 1 (+Bike)\end{tabular}}   & Fine-tuning                &0.0&0.0&0.0&-47.5&30.2&36.3       \\ 
                                                                                                & Distillation        &\textbf{65.0}&60.4&\textbf{71.5}&-95.3&25.5&28.6      \\  
                                                                                                & Det PL           &54.9&51.8&58.3&-35.9&30.9&37.0\\  
                                                                                                & COOLer       &63.8&\textbf{60.5}&71.2&\textbf{11.3}&\textbf{33.1}&\textbf{42.2}     \\ 
                                                                                                & \cellcolor{lightgray}Oracle       &\cellcolor{lightgray}64.8&\cellcolor{lightgray}60.8&\cellcolor{lightgray}71.9&\cellcolor{lightgray}24.9&\cellcolor{lightgray}36.3& \cellcolor{lightgray}47.6  \\ \midrule
\multirow{5}{*}{\begin{tabular}[c]{@{}l@{}}\textbf{V}$\rightarrow$\textbf{B}$\rightarrow$\textbf{H}\\ Stage 2 (+Human)\end{tabular}}  & Fine-tuning             &0.0&0.0&0.0&32.9&\textbf{43.2}&55.3 \\  
                                                                                                & Distillation          &61.3&58.8&69.4&21.5&39.4&50.6    \\  
                                                                                                & Det PL      &54.9&52.8&59.3&34.2&43.0&55.2\\  
                                                                                                & COOLer   &\textbf{61.7}&\textbf{59.8}&\textbf{70.7}&\textbf{43.1}&43.1&\textbf{55.9}     \\ 
                                                                                                 &\cellcolor{lightgray}Oracle    &\cellcolor{lightgray}64.3&\cellcolor{lightgray}60.4&\cellcolor{lightgray}71.6&\cellcolor{lightgray}45.9&\cellcolor{lightgray}44.3&\cellcolor{lightgray}57.5     \\ \bottomrule \bottomrule 
\end{tabular}

\end{subtable}
\end{table*}
\begin{table*}[htbp]
\scriptsize
\setlength\tabcolsep{1.6pt}
\renewcommand\arraystretch{0.6}
\caption{ Performance of old and new classes on \shift.  We conduct experiments on \textbf{M}$\rightarrow$\textbf{L}, \textbf{G}$\rightarrow$\textbf{S} and \textbf{V}$\rightarrow$\textbf{B}$\rightarrow$\textbf{H}  settings.
We compare COOLer with the Fine-tuning and Det PL baselines and the oracle tracker.}
\label{tab:shift_exp_full}
\begin{subtable}[htbp]{1.0\textwidth} 
\caption{Mean Metrics}
\centering
\label{tab:shift_exp_full_a}
\begin{tabular}{@{}llcccc|cccc@{}}

\toprule \toprule
\multirow{2}{*}{\begin{tabular}[c]{@{}l@{}}\textbf{Setting}\\ Stage (+New Classes)\end{tabular}} & &   \multicolumn{4}{c}{Old Classes}                                               & \multicolumn{4}{c}{New Classes}                                               \\ \cmidrule(l){3-10} 
 & Method                                                                         & \tiny{mMOTA}          & \tiny{mHOTA }         & \tiny{mIDF1}          & \tiny{mAP}             & \tiny{mMOTA}          &  \tiny{mHOTA}          & \tiny{mIDF1}          & \tiny{mAP}            \\ \midrule \midrule

\textbf{M}$\rightarrow$\textbf{L}  Stage 0 (Pedestrian) &      & - & - &- & - & 53.7   & 46.1  &  54.4 &  43.0\\
\midrule
\multirow{5}{*}{\begin{tabular}[c]{@{}l@{}}\textbf{M}$\rightarrow$\textbf{L}\\ Stage 1 (+Car) \end{tabular}}     & Fine-tuning        & 0.0           & 0.0           & 0.0           & 0.0        &   \textbf{51.7}      &    \textbf{57.0}      & \textbf{61.8}    &  \textbf{50.6}         \\ 
                                                                                                & Det PL               &  38.5        & 37.2       &   38.3       &  41.6          &  50.8       &  56.4         &  61.0         &   49.6               \\  
                                                                                                & COOLer       & \textbf{51.1} &  \textbf{45.5} & \textbf{53.1} &  \textbf{41.7}  & 50.6 & 56.4  &  60.8        &  49.7          \\ 
                                                                                               &  \cellcolor{lightgray}Oracle         
                                                                                               &\cellcolor{lightgray}53.7& \cellcolor{lightgray}46.1 & \cellcolor{lightgray}55.1 & \cellcolor{lightgray}41.8 & \cellcolor{lightgray}53.7 &  \cellcolor{lightgray}57.4& \cellcolor{lightgray}62.2 & \cellcolor{lightgray}50.6\\\midrule 
\multirow{5}{*}{\begin{tabular}[c]{@{}l@{}}\textbf{M}$\rightarrow$\textbf{L}\\ Stage 2 (+Truck)\end{tabular}}          & Fine-tuning          & 0.0           & 0.0           & 0.0           & 0.0            &     35.0    &  53.9      &  58.7         &  \textbf{46.4}               \\ 
                                                                                                & Det PL         &   34.5      &  40.3         &    39.9     & 44.4  &  34.7    & 53.7     &   58.0        &    45.7                 \\  
                                                                                                & COOLer     & \textbf{49.2} &  \textbf{50.3} &  \textbf{56.3}  & \textbf{45.2}  & \textbf{37.2}  &\textbf{ 54.0}   & \textbf{59.4}  &  44.1       \\ 
                                                                                                 & \cellcolor{lightgray}Oracle        & \cellcolor{lightgray}53.7 & \cellcolor{lightgray}51.8& \cellcolor{lightgray}58.7& \cellcolor{lightgray}46.2  & \cellcolor{lightgray}50.5 & \cellcolor{lightgray}57.0 & \cellcolor{lightgray}64.9 & \cellcolor{lightgray}45.6        \\ \midrule
  \textbf{G}$\rightarrow$\textbf{S}  Stage 0 (General)  &     & - & - &- & - & 50.8   & 53.1  & 60.0 &  46.0\\ \midrule
\multirow{5}{*}{\begin{tabular}[c]{@{}l@{}}\textbf{G}$\rightarrow$\textbf{S}\\ Stage 1 (+Specific)\end{tabular}} & Fine-tuning            & 0.0           & 0.0           & 0.0           & 0.0                   & 38.9          & 48.2          & 53.9          &     40.1       \\  
                                                                                               & Det PL       &   49.2          & 50.0          &   54.6          &   44.7       & \textbf{42.3} &  \textbf{49.6} &  \textbf{56.2} & \textbf{41.7}  \\  
                                                                                               & COOLer          & \textbf{50.0 }& \textbf{52.9} & \textbf{59.7} &   \textbf{44.9}        & 42.1          & 48.7          & 54.4          &  40.5         \\ 
                                                                                                &\cellcolor{lightgray}Oracle        &\cellcolor{lightgray}52.6        & \cellcolor{lightgray}53.5        & \cellcolor{lightgray}60.7          & \cellcolor{lightgray}46.0        &  \cellcolor{lightgray}45.0 & \cellcolor{lightgray}48.7  & \cellcolor{lightgray}54.2  &  \cellcolor{lightgray}41.6\\  
                                                                                               \midrule \midrule
                                                                                                \textbf{V}$\rightarrow$\textbf{B}$\rightarrow$\textbf{H}  Stage 0 (Vehicle) &  & - & - &- & - &  47.2   &   52.1&  57.4&  45.2 \\ \midrule
\multirow{5}{*}{\begin{tabular}[c]{@{}l@{}}\textbf{V}$\rightarrow$\textbf{B}$\rightarrow$\textbf{H}\\ Stage 1 (+Bike)\end{tabular}}   & Fine-tuning               & 0.0           & 0.0           & 0.0           & 0.0         & 40.2          &  \textbf{51.0} & 56.1          &  \textbf{41.4}      \\  
                                                                                               & Det PL                      & 39.1          & 44.7          & 46.8          &        43.5          & 39.8          & 50.7          & \textbf{57.3} & 39.8  \\  
                                                                                               & COOLer  & \textbf{47.0} & \textbf{49.7} & \textbf{58.2} &  \textbf{44.9}          & \textbf{40.7} & 49.7          & 56.3          &           38.3
                                                                                               \\ 
                                                                                                &\cellcolor{lightgray}Oracle      & \cellcolor{lightgray}48.9          & \cellcolor{lightgray}52.0        &  \cellcolor{lightgray}57.7        &   \cellcolor{lightgray}45.6     & \cellcolor{lightgray}46.3  & \cellcolor{lightgray}52.4  &\cellcolor{lightgray}58.4   & \cellcolor{lightgray}42.1 \\  \midrule
\multirow{5}{*}{\begin{tabular}[c]{@{}l@{}}\textbf{V}$\rightarrow$\textbf{B}$\rightarrow$\textbf{H}\\ Stage 2  (+Human)\end{tabular}}  & Fine-tuning               & 0.0           & 0.0           & 0.0           & 0.0        & \textbf{ 53.4} &  \textbf{46.0} & \textbf{ 54.6} & \textbf{42.9}  \\  
                                                                                               & Det PL               & 34.2          & 40.6          & 41.7          &  41.6              & 53.0          & 45.5          & 53.9          &   42.1        \\  
                                                                                               & COOLer     &  \textbf{45.8} & \textbf{51.7}  & \textbf{58.3} &    \textbf{42.4}       & 52.9          & 44.9          & 53.4          &      41.4     \\ 
                                                                                                &\cellcolor{lightgray}Oracle      & \cellcolor{lightgray} 47.8       & \cellcolor{lightgray}52.1         &\cellcolor{lightgray}58.0           & \cellcolor{lightgray}44.2         & \cellcolor{lightgray}53.7  &\cellcolor{lightgray}46.1  & \cellcolor{lightgray}55.1 & \cellcolor{lightgray}41.8 \\  \bottomrule \bottomrule
\end{tabular}
\end{subtable}

\begin{subtable}[htbp]{1.0\textwidth} 
\caption{Overall Metrics}
\setlength\tabcolsep{5.3pt}
\centering
\label{tab:shift_exp_full_b}
\begin{tabular}{@{}llccc|ccc@{}}
\toprule \toprule
\multirow{2}{*}{\begin{tabular}[c]{@{}l@{}}\textbf{Setting}\\ Stage (+New Classes)\end{tabular}} & &   \multicolumn{3}{c}{Old Classes}                                               & \multicolumn{3}{c}{New Classes}                                               \\ \cmidrule(l){3-8} 
 & Method                                                                         & MOTA         & HOTA          & IDF1       & MOTA         & HOTA          & IDF1              \\ \midrule \midrule

\textbf{M}$\rightarrow$\textbf{L}  Stage 0 (Pedestrian) &       & - &- & - &53.7  &46.1& 54.4\\
\midrule
\multirow{5}{*}{\begin{tabular}[c]{@{}l@{}}\textbf{M}$\rightarrow$\textbf{L}\\ Stage 1 (+Car) \end{tabular}}     & Fine-tuning     &0.0&0.0&0.0&\textbf{51.7}&\textbf{57.0}&\textbf{61.8}     \\ 
                                                                                                & Det PL          &38.5&37.2&38.3&50.8&56.4&61.0            \\  
                                                                                                & COOLer     &\textbf{51.1}&\textbf{45.5}&\textbf{53.1}&50.6&56.4&60.8          \\ 
                                                                                               &\cellcolor{lightgray}Oracle         
                                                                                            &\cellcolor{lightgray}53.7&\cellcolor{lightgray}46.1&\cellcolor{lightgray}55.1&\cellcolor{lightgray}53.7&\cellcolor{lightgray}57.4&\cellcolor{lightgray}62.2\\\midrule 
\multirow{5}{*}{\begin{tabular}[c]{@{}l@{}}\textbf{M}$\rightarrow$\textbf{L}\\ Stage 2 (+Truck)\end{tabular}}          & Fine-tuning        &0.0&0.0&0.0&35.0&53.9&58.7           \\ 
                                                                                                & Det PL       &33.4&41.8&38.8&34.7&53.7&58.0         \\  
                                                                                                & COOLer &\textbf{49.1}&\textbf{50.3}&\textbf{56.1}&\textbf{37.2}&\textbf{54.0}&\textbf{59.4}     \\ 
                                                                                                 & \cellcolor{lightgray}Oracle    &\cellcolor{lightgray}53.7&\cellcolor{lightgray}51.4&\cellcolor{lightgray}58.4&\cellcolor{lightgray}50.5&\cellcolor{lightgray}57.0&\cellcolor{lightgray}64.9       \\ \midrule
  \textbf{G}$\rightarrow$\textbf{S}  Stage 0 (General)  &      & - &- & - & 50.8  &53.1&60.0\\ \midrule
\multirow{5}{*}{\begin{tabular}[c]{@{}l@{}}\textbf{G}$\rightarrow$\textbf{S}\\ Stage 1 (+Specific)\end{tabular}} & Fine-tuning          &0.0&0.0&0.0&39.2&48.6&54.2 \\  
                                                                                               & Det PL    &  49.5&48.8&52.2  &\textbf{42.5}&\textbf{50.1}&\textbf{56.5}\\  
                                                                                             & COOLer         &\textbf{51.9}&\textbf{51.7}&\textbf{57.9}&42.4&49.0&54.7    \\ 
                                                                                                &\cellcolor{lightgray}Oracle      &\cellcolor{lightgray}53.4&\cellcolor{lightgray}51.8&\cellcolor{lightgray}58.9&\cellcolor{lightgray}45.3&\cellcolor{lightgray}48.9&\cellcolor{lightgray}54.4\\  
                                                                                               \midrule \midrule
                                                                                                \textbf{V}$\rightarrow$\textbf{B}$\rightarrow$\textbf{H}  Stage 0 (Vehicle) &  & - &- & - & 47.2  &52.1& 57.4\\ \midrule
\multirow{5}{*}{\begin{tabular}[c]{@{}l@{}}\textbf{V}$\rightarrow$\textbf{B}$\rightarrow$\textbf{H}\\ Stage 1 (+Bike)\end{tabular}}   & Fine-tuning              &0.0&0.0&0.0&40.7&\textbf{51.2}&56.4      \\  
                                                                                               & Det PL                &41.6&46.7&47.1&40.3&51.0&\textbf{57.7} \\  
                                                                                               & COOLer  &\textbf{50.9}&\textbf{56.0}&\textbf{61.0}&\textbf{41.3}&50.1&56.7
                                                                                               \\ 
                                                                                                &\cellcolor{lightgray}Oracle    &\cellcolor{lightgray}52.3&\cellcolor{lightgray}56.0&\cellcolor{lightgray}61.2&\cellcolor{lightgray}46.6&\cellcolor{lightgray}52.5&\cellcolor{lightgray}58.5\\  \midrule
\multirow{5}{*}{\begin{tabular}[c]{@{}l@{}}\textbf{V}$\rightarrow$\textbf{B}$\rightarrow$\textbf{H}\\ Stage 2  (+Human)\end{tabular}}  & Fine-tuning              &0.0&0.0&0.0&\textbf{53.4}&\textbf{46.0}&\textbf{54.6}  \\  
                                                                                               & Det PL           &29.7&36.7&34.0&53.0&45.5&53.9     \\  
                                                                                               & COOLer   &\textbf{49.1}&\textbf{55.2}&\textbf{60.4}&52.9&44.9&53.4  \\ 
                                                                                                &\cellcolor{lightgray}Oracle   &\cellcolor{lightgray}51.5&\cellcolor{lightgray}55.5&\cellcolor{lightgray}60.8&\cellcolor{lightgray}53.7&\cellcolor{lightgray}46.1&\cellcolor{lightgray}55.1\\  \bottomrule \bottomrule
\end{tabular}
\end{subtable}
\end{table*}
\begin{table*}[htbp]
\scriptsize
\setlength\tabcolsep{1.7pt}
\caption{Performance of old and new classes in the ablation study on method components.  We ablate on the choice of pseudo-labels (PL) and contrastive (CT) loss for COOLer on \textbf{M}$\rightarrow$\textbf{L} setting on BDD100k. We compare training with our proposed  pseudo-labels generated by the tracker (Track) and the pseudo-labels by the detector (Det). We also compare our proposed class-incremental contrastive loss (Ours) with the contrastive loss proposed in OWOD~\cite{joseph2021towards}.}
\label{tab:ab_full}
\begin{subtable}[htbp]{1.0\textwidth}
\caption{Mean Metrics}
\centering
\label{tab:ab_full_a}
\begin{tabular}{@{}lcccccc|cccc@{}}
\toprule
 
\multirow{2}{*}{\begin{tabular}[c]{@{}l@{}}\textbf{Setting}\\ Stage (+New Classes)\end{tabular}}                                        & \multicolumn{2}{c}{Components}   & \multicolumn{4}{c}{Old Classes} & \multicolumn{4}{c}{New Classes} \\\cmidrule(l){2-3} \cmidrule(l){4-11}                                                                         & PL & CT Loss                      & \tiny{mMOTA}          & \tiny{mHOTA}          & \tiny{mIDF1}          & \tiny{mAP}     &\tiny{mMOTA}          &  \tiny{mHOTA}          & \tiny{mIDF1}          & \tiny{mAP}        \\ \midrule  
\multirow{4}{*}{\begin{tabular}[c]{@{}l@{}}\textbf{M}$\rightarrow$\textbf{L}\\ Stage 1 \\ (+Pedestrian)\end{tabular}} & Det      & \xmark                              &  60.1        & 55.7          &    63.0       &  55.1          & 33.3         &  43.4         & 55.4          &  \textbf{38.4}   \\  
                                                                                            & Track      & \xmark                         &   65.8 & 61.6 & 72.5 & 57.0 & \textbf{42.5} & \textbf{43.7} & \textbf{56.6} & 37.5\\  
                                                                                            & Track      & OWOD~\cite{joseph2021towards}    & 65.4 & 61.3 & 72.1 & \textbf{57.1} & 42.0 & 43.1 &55.6 & 37.3   \\   
                                                                                            & Track      & Ours                                  & \textbf{66.0} & \textbf{61.8} & \textbf{72.8} & \textbf{57.1}   & \textbf{42.5} & 43.4 &  55.8        & 37.7  \\ \midrule
\multirow{4}{*}{\begin{tabular}[c]{@{}l@{}}\textbf{M}$\rightarrow$\textbf{L}\\ Stage 2 \\ (+Truck)\end{tabular}}      & Det      & \xmark                    &  49.1        & 49.3          &  58.8        &  45.3         &  6.4             &  42.9         & 49.3          &  \textbf{36.9}       \\  
                                                                                            & Track     & \xmark                         & 50.3 & 51.6 &  62.9 &  \textbf{46.1}& \textbf{29.5} & \textbf{44.3} & \textbf{52.7} & 36.3        \\  
                                                                                            & Track      & OWOD~\cite{joseph2021towards}    & 49.1 & 51.2 & 62.6 & 45.9 & 27.6 & 43.8 & 51.7 & 36.7\\  
                                                                                            & Track      & Ours                   & \textbf{50.8} &  \textbf{51.8}& \textbf{63.3} & \textbf{46.1} & 26.7 & 43.9 & 52.1  & 35.7 \\ \midrule
\multirow{4}{*}{\begin{tabular}[c]{@{}l@{}}\textbf{M}$\rightarrow$\textbf{L}\\ Stage 3 \\ (+Bus)\end{tabular}}        & Det      & \xmark                  & 36.5           & 44.1           &     51.2      &   40.8         &   -53.2        &  40.0          & 42.8          & 41.3  \\\ 
                                                                                            & Track      & \xmark                        & 40.4 & 48.1 & 58.3& \textbf{41.8} & 10.2 & 45.8 & 52.8 & 42.0    \\  
                                                                                            & Track      & OWOD~\cite{joseph2021towards}  & 39.2 &  47.9 & 58.0 &  \textbf{41.8} & \textbf{14.6} & \textbf{46.2} & \textbf{53.8} & 42.2\\  
                                                                                            & Track      & Ours                     &  \textbf{41.2} & \textbf{48.5} &  \textbf{58.6} & 41.7  & 12.4  &  \textbf{46.2} &  53.3 &  \textbf{42.4} \\ \midrule
\multirow{4}{*}{\begin{tabular}[c]{@{}l@{}}\textbf{M}$\rightarrow$\textbf{L}\\ Stage 4 \\ (+Bicycle)\end{tabular}}    & Det      & \xmark          &  18.8         &  39.6          &  44.4          &  37.9         &  -57.6       &  29.7          &    34.5       &  \textbf{20.3} \\  
                                                                                            & Track     & \xmark                    & 29.1 & 46.6 & 55.8 & 40.4 & 11.9 & 32.8 & 41.4 & 18.8 \\   
                                                                                            & Track   & OWOD~\cite{joseph2021towards}   & 28.3 &  46.3 & 55.3 & 36.1 & 10.5 & 33.3 & 42.1 & 18.6         \\  
                                                                                            & Track   & Ours                      & \textbf{32.2} & \textbf{47.1}  & \textbf{56.7}  & \textbf{40.7}  & \textbf{14.1}  & \textbf{33.5}  & \textbf{42.6}  & 19.6     \\ \bottomrule

\end{tabular}
\end{subtable}
\begin{subtable}[htbp]{1.0\textwidth}
\caption{Overall Metrics}
\centering
\setlength\tabcolsep{4.9pt}
\label{tab:ab_full_b}
\begin{tabular}{@{}lccccc|ccc@{}}
\toprule
 
\multirow{2}{*}{\begin{tabular}[c]{@{}l@{}}\textbf{Setting}\\ Stage (+New Classes)\end{tabular}}                                        & \multicolumn{2}{c}{Components}   & \multicolumn{3}{c}{Old Classes} & \multicolumn{3}{c}{New Classes} \\\cmidrule(l){2-3} \cmidrule(l){4-9}                                                                         & PL & CT Loss                      &  MOTA         & HOTA          & IDF1    &  MOTA         & HOTA          & IDF1                   \\ \midrule  
\multirow{4}{*}{\begin{tabular}[c]{@{}l@{}}\textbf{M}$\rightarrow$\textbf{L}\\ Stage 1 \\ (+Pedestrian)\end{tabular}} & Det      & \xmark                  &60.1&55.7&63.0&33.3&43.4&55.4  \\  
                                                                                            & Track      & \xmark                    &65.8&61.6&72.5&\textbf{42.5}&\textbf{43.7}&\textbf{56.6}\\  
                                                                                            & Track      & OWOD~\cite{joseph2021towards}    &65.4&61.3&72.1&42.0&43.1&55.6   \\   
                                                                                            & Track      & Ours                  &\textbf{66.0}&\textbf{61.8}&\textbf{72.8}&\textbf{42.5}&43.4&55.8 \\ \midrule
\multirow{4}{*}{\begin{tabular}[c]{@{}l@{}}\textbf{M}$\rightarrow$\textbf{L}\\ Stage 2 \\ (+Truck)\end{tabular}}      & Det      & \xmark            &60.8&56.5&66.3&6.4&42.9&49.3      \\  
                                                                                            & Track     & \xmark      &60.2&58.3&69.2&\textbf{29.5}&\textbf{44.3}&\textbf{52.7}    \\  
                                                                                            & Track      & OWOD~\cite{joseph2021towards}   &59.6&58.1&69.0&27.6&43.8&51.7\\  
                                                                                            & Track      & Ours            &\textbf{60.8}&\textbf{58.7}&\textbf{69.6}&26.7&43.9&52.1 \\ \midrule
\multirow{4}{*}{\begin{tabular}[c]{@{}l@{}}\textbf{M}$\rightarrow$\textbf{L}\\ Stage 3 \\ (+Bus)\end{tabular}}        & Det      & \xmark           &55.9&53.9&62.1&-53.2&40.0&42.8  \\\ 
                                                                                            & Track      & \xmark                   &55.4&56.5&67.1&10.2&45.8&52.8 \\  
                                                                                            & Track      & OWOD~\cite{joseph2021towards}  &54.7&56.3&66.8&\textbf{14.6}&\textbf{46.2}&\textbf{53.8}\\  
                                                                                            & Track      & Ours         &\textbf{56.8}&\textbf{57.1}&\textbf{67.8}&12.4&\textbf{46.2}&53.3 \\ \midrule
\multirow{4}{*}{\begin{tabular}[c]{@{}l@{}}\textbf{M}$\rightarrow$\textbf{L}\\ Stage 4 \\ (+Bicycle)\end{tabular}}    & Det     & \xmark   &41.2&43.2&46.2&-57.6&29.7&34.5 \\  
                                                                                            & Track     & \xmark        &51.2&55.3&65.6&11.9&32.8&41.1 \\   
                                                                                            & Track   & OWOD~\cite{joseph2021towards} &50.1&55.0&65.2&10.5&33.3&42.1      \\  
                                                                                            & Track   & Ours        &\textbf{53.5}&\textbf{55.9}&\textbf{66.3}&\textbf{14.1}&\textbf{33.5}&\textbf{42.6}   \\ \bottomrule

\end{tabular}
\end{subtable}
\end{table*}
\newpage

\section{Class-Incremental Instance Representation Learning}
We perform further analysis on our class-incremental instance representation learning with the proposed contrastive loss. We first compare the position of the contrastive loss by applying it either on the detection head similar to~\cite{joseph2021towards} or the similarity head. As shown in Tab.~\ref{tab:position_ct_loss}, the performance of the instance representation learning does not vary much with its position, while applying it on the similarity head has a slight improvement especially in the overall metrics.

We perform sensitivity analysis on two important parameters for our contrastive loss, the standard deviation vector of the prior distribution $\boldsymbol{\sigma}_p$ and the hinge factor for the pushing loss $\Delta_{\mathrm{push}}$. The results are shown in Tab.~\ref{tab:sigma_p} and Tab.~\ref{tab:delta_push}, respectively. We observe that varying these parameters won't significantly affect the performance in the initial incremental stages with only a few classes learnt. However,  setting $\boldsymbol{\sigma}_p$ to be too small may hurt the tracking performance. This would imply that while clustering the class instance representations is beneficial to detection, it is still important to maintain  intra-class variability and discrimination properties within the class embedding distribution for data association. Also, setting $\Delta_{\mathrm{push}}$ to be too large would hurt both detection and tracking performance in later incremental stages, as distributions of different classes may overlap with each other given the limited embedding space size.
\begin{table}[htbp]
\centering
\scriptsize
\setlength\tabcolsep{2pt}
\caption{Results for varying the position of the class-incremental contrastive learning on the \textbf{M}$\rightarrow$\textbf{L} setting on \bdd.}
\label{tab:position_ct_loss}
\begin{tabular}{@{}lcccccccc@{}}
\toprule
 \multirow{2}{*}{\begin{tabular}[c]{@{}l@{}}\textbf{Setting}\\ Stage (+New Classes) \end{tabular}}       & & \multicolumn{7}{c}{All Classes}                    \\ \cmidrule(l){3-9}     
                                                                                    &       Position                                           & mMOTA          & mHOTA          & mIDF1  & MOTA & HOTA & IDF1        & mAP         \\ \midrule
\multirow{2}{*}{\begin{tabular}[c]{@{}l@{}}\textbf{M$\rightarrow$L}\\ Stage 1 (+Pedestrian) \end{tabular}} & Detection                         & 54.3          & 52.6          & 64.6         &62.2&59.1&70.2&    46.8\\  
                                                                                            &  Similarity  & 54.2          & 52.6          &  64.3 &62.7&59.5&70.5&   47.4    \\  
                                                                                       \midrule
\multirow{2}{*}{\begin{tabular}[c]{@{}l@{}}\textbf{M$\rightarrow$L}\\ Stage 2 (+Truck)\end{tabular}}      & Detection           & 43.4          & 48.8          & 59.2          &58.0&57.3&68.0&       42.1                                     \\ 
                                                                                            & Similarity              & 42.8 & 49.2          & 59.6         &58.6&57.9&68.7 &  42.6 \\  
                                                                                              \midrule
\multirow{2}{*}{\begin{tabular}[c]{@{}l@{}}\textbf{M$\rightarrow$L}\\  Stage 3 (+Bus) \end{tabular}}        & Detection             & 35.1          & 47.5          & 57.4         &54.5&56.1&66.8 &         40.8       \\  
                                                                                            &  Similarity              & 34.0          & 47.9          & 57.3         &55.8&56.8&67.4 &  41.9     \\  
                                                                                             \midrule
\multirow{2}{*}{\begin{tabular}[c]{@{}l@{}}\textbf{M$\rightarrow$L}\\ Stage 4 (+Bicycle)\end{tabular}}    
                                                                                            & Detection  &  28.6 & 43.9 & 53.8 & 51.2 &55.1&65.7&     35.0    \\  
                                                                                            & Similarity                            & 28.6          & 44.4          & 53.9         &53.2&55.7&66.1 &  36.5  \\ \bottomrule
\end{tabular}

\end{table}
\begin{table}[htbp]
\centering
\scriptsize
\setlength\tabcolsep{2pt}
\renewcommand\arraystretch{0.8}
\caption{Result for varying the standard deviation vector of the prior distribution $\boldsymbol{\sigma}_{p}$ on the \textbf{M}$\rightarrow$\textbf{L} setting on \bdd. }
\label{tab:sigma_p}
\begin{tabular}{@{}lcccccccc@{}}
\toprule
 \multirow{2}{*}{\begin{tabular}[c]{@{}l@{}}\textbf{Setting}\\ Stage (+New Classes)\end{tabular}}       & & \multicolumn{7}{c}{All Classes}                    \\ \cmidrule(l){3-9}     
                                                                                    &       $\boldsymbol{\sigma}_p$                                    &  mMOTA          & mHOTA          & mIDF1  & MOTA & HOTA & IDF1        & mAP          \\ \midrule
\multirow{3}{*}{\begin{tabular}[c]{@{}l@{}}\textbf{M$\rightarrow$L}\\ Stage 1 (+Pedestrian) \end{tabular}} & 0.01$\cdot\Vec{\mathbf{1}}$                        &  54.2          & 52.7          &  64.4         & 62.6&59.5&70.5 &  47.4  \\  
                                                                                            &  0.05$\cdot\Vec{\mathbf{1}}$  & 54.2          & 52.6          &  64.3 &62.7 &59.5&70.5&   47.4    \\  
                                                                                            &  0.10$\cdot\Vec{\mathbf{1}}$    & 54.5          & 52.8          &  64.5 &62.7 &59.6&70.7&  47.3     \\  
                                                                                       \midrule
\multirow{3}{*}{\begin{tabular}[c]{@{}l@{}}\textbf{M$\rightarrow$L}\\ Stage 2 (+Truck) \end{tabular}}      & 0.01$\cdot\Vec{\mathbf{1}}$          &  43.1          & 49.2         & 59.6         & 58.6&57.8&68.8&  42.7                           \\ 
                                                                                            & 0.05$\cdot\Vec{\mathbf{1}}$             & 42.8 & 49.2          & 59.6         &58.6&57.9&68.7 &  42.6\\  & 0.10$\cdot\Vec{\mathbf{1}}$           & 44.3 & 49.3          & 59.8          &59.3 &58.1&68.9&    42.5 \\  
                                                                                              \midrule
\multirow{3}{*}{\begin{tabular}[c]{@{}l@{}}\textbf{M$\rightarrow$L}\\  Stage 3 (+Bus) \end{tabular}}        & 0.01$\cdot\Vec{\mathbf{1}}$          & 32.7          & 47.6          & 55.3         &56.3&56.7&67.4& 41.8                           \\ 
                                                                                            &  0.05$\cdot\Vec{\mathbf{1}}$               & 34.0          & 47.9          & 57.3          &55.8&56.8&67.4&     41.9   \\ 
                                                                                              &  0.10$\cdot\Vec{\mathbf{1}}$              & 34.8          & 47.9          & 57.4          &57.3&56.9&67.6&   41.7  \\ 
                                                                                             \midrule
\multirow{3}{*}{\begin{tabular}[c]{@{}l@{}}\textbf{M$\rightarrow$L}\\ Stage 4 (+Bicycle) \end{tabular}}    
                                                                                            & 0.01$\cdot\Vec{\mathbf{1}}$   &  28.7 & 44.0 & 53.3 &52.9 &55.8&66.4&     36.3        \\  
                                                                                            & 0.05$\cdot\Vec{\mathbf{1}}$                          & 28.6          & 44.4          & 53.9          &53.2 &55.7&66.1&  36.5    \\     & 0.10$\cdot\Vec{\mathbf{1}}$                            & 29.0          & 44.4          & 54.0          & 53.6&55.9&66.4&     36.4 \\ \bottomrule
\end{tabular}

\end{table}
\begin{table}[htbp]
\centering
\scriptsize
\setlength\tabcolsep{2pt}
\renewcommand\arraystretch{0.8}
\caption{Results for varying  the hinge factor for the pushing loss  $\Delta_\mathrm{push}$ on the \textbf{M}$\rightarrow$\textbf{L} setting on \bdd. }
\label{tab:delta_push}
\begin{tabular}{@{}lcccccccc@{}}
\toprule
 \multirow{2}{*}{\begin{tabular}[c]{@{}l@{}}\textbf{Setting}\\ Stage (+New Classes)\end{tabular}}       & & \multicolumn{7}{c}{All Classes}                    \\ \cmidrule(l){3-9}     
                                                                                    &       $\Delta_{\mathrm{push}}$                                            & mMOTA          & mHOTA          & mIDF1          & MOTA & HOTA & IDF1 &    mAP    \\ \midrule
\multirow{3}{*}{\begin{tabular}[c]{@{}l@{}}\textbf{M$\rightarrow$L}\\ Stage 1 (+Pedestrian)\end{tabular}} & 10                          & 54.4          & 52.8          & 64.4   &62.8&59.6&70.6        &  47.3 \\  
                                                                                            &  15   & 54.2          & 52.6          &  64.3 &62.7 &59.5&70.5&    47.4    \\        &  20  & 54.0          & 52.7          &  64.4 & 52.6 &59.4&70.4& 47.4      \\  
                                                                                       \midrule
\multirow{3}{*}{\begin{tabular}[c]{@{}l@{}}\textbf{M$\rightarrow$L}\\ Stage 2 (+Truck)\end{tabular}}      & 10        & 43.5          & 49.1          & 59.4         &58.8&57.8&68.7 & 42.7                               \\ 
                                                                                            & 15                  & 42.8 & 49.2          & 59.6          & 58.6&57.9& 68.7&  42.6 \\     & 20                & 43.5 & 49.1          & 59.5          &58.7 &58.0&68.8&  42.7\\  
                                                                                              \midrule
\multirow{3}{*}{\begin{tabular}[c]{@{}l@{}}\textbf{M$\rightarrow$L}\\  Stage 3 (+Bus)\end{tabular}}        & 10            &        33.8    &  47.8          &  57.5        &55.8&56.8&67.5  &  41.7                         \\  
                                                                                            &  15            & 34.0          & 47.9          & 57.3          &55.8 &56.8&67.4&  41.9   \\   &  20                & 33.2          & 47.5          & 56.9          & 55.4&56.6&67.2&     41.8\\  
                                                                                             \midrule
\multirow{3}{*}{\begin{tabular}[c]{@{}l@{}}\textbf{M$\rightarrow$L}\\ Stage 4 (+Bicycle)\end{tabular}}    
                                                                                            & 10    &28.1   & 44.4  & 53.6  &52.9&55.9&66.4 &    36.2         \\  
                                                                                            & 15                            & 28.6          & 44.4          & 53.9          &53.2&55.7&66.1&    36.5   \\  & 20                         & 28.0         &        44.0    & 53.5         &52.4&55.6&66.1  & 35.9      \\ \bottomrule
\end{tabular}

\end{table}
 
%
%

\begin{table}[htbp]
\centering
\scriptsize
\setlength\tabcolsep{2pt}
\caption{Further ablation study on method components.  We ablate on the choice of pseudo-labels (PL) and contrastive loss (CL) for COOLer on the \textbf{V}$\rightarrow$\textbf{B}$\rightarrow$\textbf{H} setting on \shift. We compare training with our proposed pseudo-labels generated by the tracker (Track) and the pseudo-labels generated by the detector (Det). We also compare our proposed contrastive loss (Ours) with the contrastive loss proposed in OWOD~\cite{joseph2021towards}.}
\label{tab:ab_shift}
\begin{tabular}{@{}lccccccccc@{}}
\toprule
\multirow{3}{*}{\begin{tabular}[c]{@{}l@{}}\textbf{Setting}\\ Stage \\(+New Classes)\end{tabular}}    & & & & & &  \\                                         & \multicolumn{2}{c}{Components} & \multicolumn{7}{c}{All Classes} \\\cmidrule(l){2-3} \cmidrule(l){4-10}                                                                               & PL & CT Loss                          & mMOTA          & mHOTA          & mIDF1          & MOTA & HOTA & IDF1 &   mAP     \\ \midrule
\multirow{4}{*}{\begin{tabular}[c]{@{}l@{}}\textbf{V}$\rightarrow$\textbf{B}$\rightarrow$\textbf{H}\\ Stage 1 \\ (+Bike)\end{tabular}} & Det      & \xmark                      & 39.4     &     47.1          &  51.0            &41.4&47.4& 48.6&   42.0        \\  
                                                                                            & Track      & \xmark                       & 44.1          & 51.0          & 57.1 &\textbf{49.6}&\textbf{55.6}&\textbf{60.9}&  42.1\\  
                                                                                            & Track      & OWOD~\cite{joseph2021towards}    & 43.8  &  51.0 & 57.4           &49.5&55.2&60.7&        42.1   \\   
                                                                                            & Track      & Ours                          & \textbf{44.5} & \textbf{51.3} & \textbf{57.5}        &49.5&55.5& \textbf{60.9}  &        \textbf{42.3}  \\ \midrule
\multirow{4}{*}{\begin{tabular}[c]{@{}l@{}}\textbf{V}$\rightarrow$\textbf{B}$\rightarrow$\textbf{H}\\ Stage 2 \\ (+Human)\end{tabular}}      & Det      & \xmark                      & 37.3 & 41.4 &43.7 &39.8&40.9&42.6&   41.7         \\  
                                                                                            & Track     & \xmark                             &  45.8 & 49.7          & 56.1          &50.2&51.0&57.2&      \textbf{42.3}     \\  
                                                                                            & Track      & OWOD~\cite{joseph2021towards}  & 45.8           & 50.4  & 56.9  &50.1&51.3&\textbf{57.8}&  42.1  \\  
                                                                                            & Track      & Ours                           & \textbf{47.0}          & \textbf{50.6} & \textbf{57.5} & \textbf{50.7}&\textbf{51.3}&57.7&   42.2 \\   \bottomrule
\end{tabular}

\end{table}
\section{Ablation Studies on \shift}
As in the main paper we conduct ablation studies on the \bdd dataset, we also conduct ablation studies on the \shift dataset for each component of COOLer in Tab.~\ref{tab:ab_shift}. We can draw the same conclusion that using the tracking pseudo-labels performs better than using the detection pseudo-labels, and applying our class-incremental instance represenation learning can further improve the performance. Our contrastive loss also performs better than the contrastive loss used in OWOD~\cite{joseph2021towards} on \shift.

\section{Full Results on \bdd Most$\rightarrow$Least Setting}
To better demonstrate the learning and forgetting of each class, we show the full results of COOLer on the \bdd Most$\rightarrow$Least setting with the performance of each class in each incremental stage in Tab.~\ref{tab:bdd_most_to_least_full}. We also compare our per-class results with the Det PL and the Distillation baseline and the oracle tracker. COOLer achieves the best performance when the class is newly learned and also alleviates catastrophic forgetting more effectively in later stages.
\begin{table*}[htbp]
\centering
\scriptsize

\caption{Per-class results for COOLer, Det PL, and the Distillation baseline on the BDD100k \textbf{M}$\rightarrow$\textbf{L} setting.}
\label{tab:bdd_most_to_least_full}
\begin{tabular}{@{}llcccc|cccc | cccc@{}}
\toprule \toprule
\multicolumn{2}{l}{\textbf{\bdd} \textbf{M$\rightarrow$L}} 
  & \multicolumn{4}{c|}{COOLer}        & \multicolumn{4}{c}{Det PL}    & \multicolumn{4}{c}{Distillation}         \\ \cmidrule(l){3-14} 
Category    & Stage         & MOTA   & HOTA & IDF1 & AP   & MOTA    & HOTA & IDF1 & AP  & MOTA    & HOTA & IDF1 & AP\\ \midrule \midrule
\multirow{8}{*}{Car}                                       & Stage 0                            & 67.6   & 62.1 & 73.3 & 58.7 
 & 67.6    & 62.1 & 73.3 & 58.7  
  & 67.6    & 62.1 & 73.3 & 58.7  
   \\
                                                           & Stage 1           & 66.0   & 61.8 & 72.8 &  57.1
                                         & 60.1    & 55.7 & 63.0 &   55.1 
                                         &68.0 & 62.1 & 73.4 & 58.3
                                       
                                        \\
                                                           & Stage 2                  & 64.8   & 61.2 & 72.1 & 55.9 
                         & 65.6    & 59.2 & 69.4 &  54.1 
                           & 66.4 & 60.8 & 72.0 & 57.7
                         \\
                                                           & Stage 3           & 63.3   & 60.4 & 71.3 & 55.3 
                                                 & 64.1    & 58.0 & 67.3 & 52.7  
                                                 &64.8 & 59.8 & 71.0 & 57.1 
                                                 \\
                                                           & Stage 4            & 61.3   & 59.5 & 70.2 & 54.9 
                                                & 47.6    & 45.0 & 47.2 & 50.8  
                                                & 61.0 & 57.4 & 68.4 & 55.9
                                              \\
                                                           & Stage 5            & 59.9   & 59.0 & 69.7 & 54.6 
                                                    & 30.8    & 35.1 & 32.7 &  50.0 
                                                      &55.2 & 53.9 & 64.4 & 54.3
                                                   \\
                                                           & Stage 6              & 57.6   & 58.3 & 68.8 & 54.0
                                                        & 43.0    & 42.6 & 44.0 & 49.0 
                                                         &47.4 & 49.1 & 58.4 & 54.1\\
                                                           & Stage 7             & 55.4   & 57.6 & 68.1 & 52.7
                                                     & 52.2    & 48.3 & 54.3 & 48.1 
                                                     & 37.2 & 47.2 & 49.4 & 53.1\\ \cmidrule(l){2-14}  & Oracle                               & 68.2   & 62.4 & 73.8 & 58.6  &$-$   &$-$&$-$&$-$  &$-$   &$-$&$-$&$-$\\ \midrule \midrule
\multirow{7}{*}{Pedestrian}                                & Stage 1                           & 42.5   & 43.4 & 55.8 &  37.7 
& 33.3    & 43.4 & 55.4 &  38.4 
& 24.9 & 41.3 & 52.6 & 36.0\\
                                                           & Stage 2               & 36.8   & 42.4 & 54.5 & 36.3 
                                                     & 32.6    & 39.3 & 48.1 &  36.4 
                                                     & 38.1 & 42.8 & 55.3 & 36.1\\
                                                           & Stage 3             & 30.4   & 41.4 & 52.9 & 35.2
                                                           & 22.6    & 32.9 & 37.4 &  35.0 
                                                           & 42.6 & 43.4 & 56.3 & 35.9\\
                                                           & Stage 4            & 25.0   & 40.5 & 51.6 & 34.9 
                                                    & 29.0    & 35.8 & 43.0 &  33.9 
                                                    & 44.2 & 41.8 & 54.9 & 34.5\\
                                                           & Stage 5                  & 21.2   & 40.0 & 51.1 &  34.1
                                                        & 34.0    & 37.2 & 46.1 &  32.6 
                                                        &39.5 & 37.5 & 48.7 & 35.0\\
                                                           & Stage 6           & 15.5   & 39.1 & 49.5 &  34.1
                                                     & 26.7    & 32.9 & 38.6 &  30.9 
                                                     & 32.2 & 33.1 & 42.2 & 33.5\\
                                                           & Stage 7         & 9.2    & 38.1 & 48.0 &  33.8
                                                            & 17.1    & 33.1 & 39.4 & 29.4 
                                                            & 24.6 & 28.3 & 34.2 & 33.1
                                                            \\ \cmidrule(l){2-14}  & Oracle                           & 46.6   & 44.7 & 57.9 &  38.0 &$-$   &$-$&$-$&$-$  &$-$   &$-$&$-$&$-$\\ \midrule \midrule
\multirow{6}{*}{Truck}                                     & Stage 2                             & 26.7   & 43.9 & 52.1 &  35.7
& 6.4     & 42.9 & 49.3 &  36.9 
& -22.4 & 39.7 & 44.6 & 34.4\\
                                                           & Stage 3             & 30.0   & 43.6 & 51.6 & 34.7  
                                                         & 22.7    & 41.3 & 49.0 &  34.6 
                                                         & 17.1 & 38.2 & 45.0 & 32.9\\
                                                           & Stage 4             & 29.8   & 43.0 & 51.5 & 33.6 
                                                     & 23.7    & 38.0 & 44.3 &  30.7 
                                                     & 24.9 & 39.0 & 46.7 & 32.3 \\
                                                           & Stage 5              & 27.3   & 42.4 & 50.8 &  31.9
                                                            & 22.0    & 34.2 & 38.8 &  28.7 
                                                            & 27.1 & 38.4 & 45.9 & 31.4\\
                                                           & Stage 6              & 26.8   & 41.2 & 49.4 &  30.7
                                                     & 22.1    & 34.6 & 39.1 &  26.8 
                                                     & 27.9 & 37.3 & 44.6 & 30.6\\
                                                           & Stage 7                   & 26.0   & 40.2 & 48.1 &  29.6
                                             & 6.2     & 31.2 & 36.5 & 21.5
                                             & 23.0 & 32.8 & 37.6 & 28.6\\ \cmidrule(l){2-14}  & Oracle                         & 34.5   & 45.4 & 54.5 & 38.3  &$-$   &$-$&$-$&$-$  &$-$   &$-$&$-$&$-$\\ \midrule \midrule
\multirow{5}{*}{Bus}                                       & Stage 3                              & 12.4   & 46.2 & 53.3 &  42.4 
& -53.2   & 62.3 & 42.8 &  41.3 
&-169.5 & 32.3 & 31.0 & 37.5 \\
                                                           & Stage 4                              & 12.7   & 45.5 & 53.6 &  39.5
                                                            & -25.0   & 39.7 & 43.0 &  36.0 
                                                            & -57.5 & 38.4 & 41.1 & 36.5
                                                            \\
                                                           & Stage 5                & 13.6   & 44.0 & 51.0 & 38.2 
                                                            & -18.8   & 38.2 & 41.2 &  33.3 
                                                            & -15.6 & 42.3 & 47.7 & 35.3\\
                                                           & Stage 6                   & 8.9    & 42.8 & 49.9 & 36.1
                                                         & -11.9   & 36.7 & 40.4 &  30.2 
                                                         & 3.1 & 43.6 & 50.0 & 34.4 \\ 
                                                           & Stage 7                & 14.0   & 43.0 & 50.6 & 35.0
                                                 & -42.6   & 34.0 & 37.0 &  25.5 
                                                 & 19.1 & 44.4 & 52.1 & 33.5\\ \cmidrule(l){2-14}  & Oracle                            & 32.1   & 47.7 & 57.3 &  42.9&$-$   &$-$&$-$&$-$  &$-$   &$-$&$-$&$-$\\ \midrule \midrule
\multirow{4}{*}{Bicycle}                                   & Stage 4                               & 14.1   & 33.5 & 42.6 & 19.6 
 & -57.6   & 29.7 & 43.0 &  20.3 
 &-169.6 & 23.7 & 24.5 & 16.3  \\
                                                           & Stage 5                  & 10.6   & 33.6 & 42.8 &  17.3
                                                         & -20.1   & 30.9 & 36.8 &  16.9  
                                                         & -31.3 & 29.8 & 35.4 & 14.8 \\
                                                           & Stage 6                 & 9.6    & 32.9 & 41.6 & 18.2 
                                                         & -0.3    & 30.1 & 37.9 &  16.2 
                                                         & 8.0 & 28.1 & 36.4 & 15.1\\
                                                           & Stage 7              & 12.4   & 31.9 & 39.6 & 17.9 
                                                         & -6.7    & 29.9 & 36.9 & 15.4  
                                                         & 13.6 & 28.5 & 36.7 &  15.7\\ \cmidrule(l){2-14}  & Oracle                        & 25.0   & 35.5 & 46.3 &  21.8 &$-$   &$-$&$-$&$-$   &$-$   &$-$&$-$&$-$\\ \midrule \midrule
\multirow{3}{*}{Rider}                                     & Stage 5                        & 23.2   & 33.7 & 44.8 & 23.1 
& -74.3   & 28.6 & 33.7 &  22.4 
& -307.7 & 22.2 & 21.9 & 19.3 \\
                                                           & Stage 6                  & 26.0   & 32.1 & 43.3 &  20.9
                                                 & -24.8   & 22.9 & 27.0 &  18.4 
                                                 &-72.9 & 27.5 & 33.2 & 18.6 \\
                                                           & Stage 7                 & 22.5   & 29.9 & 40.9 & 19.3
                                                            & -26.9   & 21.0 & 24.9 &  15.7 
                                                            &-15.7 & 31.0 & 39.6 & 18.4\\ \cmidrule(l){2-14}  & Oracle   & 30.9&35.2&48.5&    24.8                       &$-$   &$-$&$-$&$-$  &$-$   &$-$&$-$&$-$\\ \midrule \midrule
\multirow{2}{*}{Motorcycle}                                & Stage 6                      & -40.0  & 32.2 & 43.3 &  19.7 
& -513.6  & 19.0 & 15.0 &17.9 
&-1481.6 & 12.1 & 6.7 & 15.2\\
                                                           & Stage 7                              & -10.1  & 34.6 & 45.7 & 20.8
                                         & -206.3  & 23.0 & 22.4 &  16.7 
                                         & -473.1 & 19.4 & 16.3 & 14.6\\ \cmidrule(l){2-14}  & Oracle                               & 24.2   & 40.1 & 54.8 &   23.0&$-$   &$-$&$-$&$-$  &$-$   &$-$&$-$&$-$\\ \midrule \midrule
Train                                                      & Stage 7                          & -150.0 & 0.0  & 0.0  & 0.0 
  & -3194.8 & 1.3  & 0.3  & 0.0  
  & -2609.1 & 1.7 & 0.5 & 0.0 \\ \cmidrule(l){2-14}  & Oracle                             & -18.2   & 0.0 & 0.0 & 0.0  &$-$   &$-$&$-$&$-$  &$-$   &$-$&$-$&$-$\\ \bottomrule \bottomrule
\end{tabular}

\end{table*}

\section{Embedding Visualization}
\begin{figure*}[t]
\centering
\includegraphics[width=1.0\textwidth]{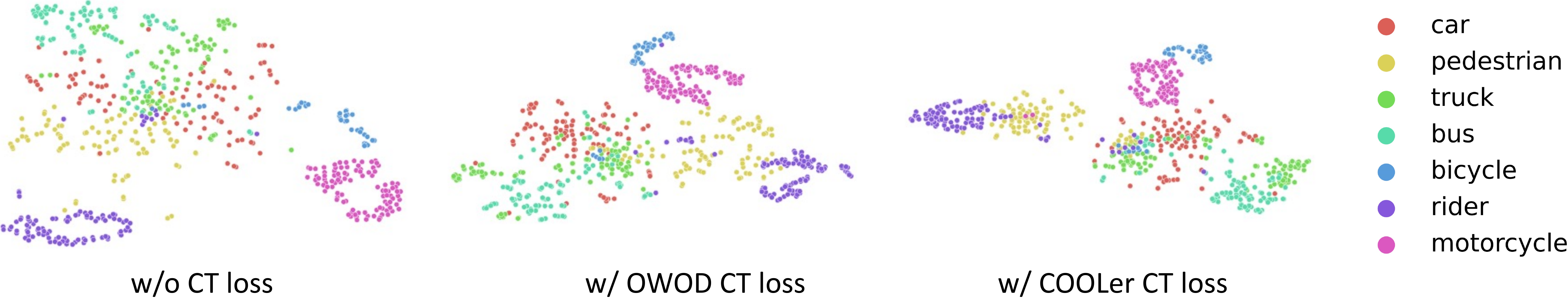} 
\caption{t-SNE visualizations of the embedding space. The embedding instances are selected from several videos in \bdd validation set. We compare COOLer's contrastive loss (CT loss) with training without CT loss and with the CT loss proposed in OWOD~\cite{joseph2021towards} in the \textbf{M}$\rightarrow$\textbf{L} setting on \bdd. We use the same color for instances of the same class.} 
\label{fig:tsne}
\end{figure*}

Fig.~\ref{fig:tsne} shows the t-SNE visualization of the embedding space for the class-incremental instance representation learning. We can see that the embedding space of training without contrastive loss is heavily entangled among different classes. Training both with the contrastive loss of OWOD~\cite{joseph2021towards} and our contrastive loss can disentangle the class embeddings, while our loss can better cluster embeddings of the same class such as cars and pedestrians. 
\end{document}